%% file: main.tex
\pdfoutput=1
\documentclass[11pt]{article}
\usepackage{acl2023}
\usepackage{times}
\usepackage{latexsym}
\usepackage{multirow}
\usepackage{sidecap}
\usepackage{amsthm}
\usepackage{amsmath}
\usepackage{amssymb}
\usepackage{amsfonts}
\usepackage{verbatim} 
\usepackage{url}
\usepackage{pifont}
\usepackage{eqparbox}
\usepackage{arydshln} 
\usepackage{bigstrut}
\usepackage{comment}
\usepackage[linewidth=1pt]{mdframed}
\usepackage{framed}
\usepackage{xcolor,colortbl}
\usepackage{pgffor}
\usepackage{comment}
\usepackage{graphicx}
\usepackage{booktabs}
\usepackage{subcaption}
\usepackage{enumitem}
\usepackage{makecell}
\usepackage{boxedminipage}
\usepackage{xspace}
\usepackage{array}
\usepackage{framed}
\usepackage{caption}
\usepackage{subcaption}
\usepackage{framed}
\usepackage{array}
\usepackage{listings}
\usepackage{adjustbox}
\usepackage{tikz}
\usetikzlibrary{tikzmark}
\usepackage{enumitem}
\usepackage{bigstrut}
\usepackage{placeins}

\usepackage[notextcomp]{stix}

\newcolumntype{L}[1]{>{\raggedright\let\newline\\\arraybackslash\hspace{0pt}}m{#1}}
\newcolumntype{C}[1]{>{\centering\let\newline\\\arraybackslash\hspace{0pt}}m{#1}}
\newcolumntype{R}[1]{>{\raggedleft\let\newline\\\arraybackslash\hspace{0pt}}m{#1}}

\usepackage[T1]{fontenc}
\usepackage[utf8]{inputenc}
\usepackage{microtype}
\usepackage{soul}

\definecolor{lightergray}{RGB}{230,230,230}
\definecolor{DarkRed}{RGB}{130,25,0}
\definecolor{DarkGreen}{RGB}{30,130,30}
\definecolor{DarkBlue}{RGB}{0,0,250}
\definecolor{purple}{rgb}{0.5,0,1}
\definecolor{dcyan}{rgb}{0.2,0.6,0.5}
\definecolor{darkgreen}{rgb}{0,200,0}
\definecolor{light-gray}{gray}{0.95} 
\definecolor{darkred}{RGB}{200,0,0}
\definecolor{lightgreen}{RGB}{231,255,219}
\definecolor{lightred}{RGB}{252,231,234}
\definecolor{lightyellow}{RGB}{250,253,191}

\newcommand{\greentext}[1]{\colorbox{lightgreen}{#1}\xspace}

\newcommand{\cmark}{\textcolor{DarkGreen}{\ding{51}}}
\newcommand{\xmark}{\textcolor{red}{\ding{55}}}%
\newcommand{\todo}[1]{{\color{red} [TODO: {#1}]}}

\newcommand{\name}{\textsc{Self-Instruct}}

\newcommand{\supernat}{\textsc{Super-NaturalInstructions}}
\newcommand{\supernatShort}{\textsc{SuperNI}}
\newcommand{\tkinstruct}{\textsc{T$k$-Instruct}}
\newcommand{\tzero}{\textsc{T$0$}}

\newcommand{\gptthree}{\textsc{GPT3}}
\newcommand{\gptself}{\textsc{GPT3}$_{\textsc{Self-Inst}}$}
\newcommand{\gptinstruct}[1]{$\text{InstructGPT}_{\text{#1}}$}

\newcommand{\promptsource}{\textsc{PromptSource}}

\title{
\vspace*{-0.5in}
{{\small \hfill ACL 2023}\\
\vspace*{.25in}} 
\name{}: Aligning Language Models \\ with Self-Generated Instructions}

\author{
Yizhong Wang\textsuperscript{$\clubsuit$} \;\;\;  
Yeganeh Kordi\textsuperscript{$\diamondsuit$} \;\;\;
Swaroop Mishra\textsuperscript{$\heartsuit$} \;\;\; 
\textbf{Alisa Liu}\textsuperscript{$\clubsuit$}\\ 
\textbf{Noah A. Smith}\textsuperscript{$\clubsuit$}\textsuperscript{$+$} \;\;\;
\textbf{Daniel Khashabi}\textsuperscript{$\spadesuit$} \; \;\;
\textbf{Hannaneh Hajishirzi}\textsuperscript{$\clubsuit$}\textsuperscript{$+$}\\
  \textsuperscript{$\clubsuit$}University of Washington \; 
  \textsuperscript{$\diamondsuit$}Tehran Polytechnic \;
  \textsuperscript{$\heartsuit$}Arizona State University \\
  \textsuperscript{$\spadesuit$}Johns Hopkins University\; \textsuperscript{$+$}Allen Institute for AI \\
  {  \texttt{yizhongw@cs.washington.edu} } \\}

\begin{document}
\maketitle
\begin{abstract}

Large ``instruction-tuned'' language models (i.e., finetuned to respond to instructions) have demonstrated a remarkable ability to generalize zero-shot to new tasks. 
Nevertheless, they depend heavily on human-written instruction data that is often limited in quantity, diversity, and creativity, therefore hindering the generality of the tuned model. 
We introduce \name{}, a  framework for improving the instruction-following capabilities of pretrained language models by bootstrapping off their own generations. 
Our pipeline generates instructions, input, and output samples from a language model, then filters invalid or similar ones before using them to finetune the original model.
Applying our method to the vanilla \gptthree{}, we demonstrate a 33\% absolute improvement over the original model on \supernat{}, on par with the performance of \gptinstruct{001},\footnote{\label{footnote:gpt:instruct}Unless otherwise specified, our comparisons are with the {\tt text-davinci-001} engine. 
We focus on this engine since it is the closest to our experimental setup: supervised finetuning with human demonstrations. The newer engines are more powerful, though they use more data (e.g., code completion or latest user queries) or algorithms (e.g., PPO) that are difficult to compare with.} which was trained with private user data and human annotations.
For further evaluation, we curate a set of expert-written instructions for novel tasks, and show through human evaluation that tuning GPT3 with \name{} outperforms using existing public instruction datasets by a large margin, leaving only a 5\% absolute gap behind \gptinstruct{001}.
\name{} provides an almost annotation-free method for aligning pretrained language models with instructions, and we release our large synthetic dataset to facilitate future studies on instruction tuning.\footnote{Code and data are available at \url{https://github.com/yizhongw/self-instruct}
}
\end{abstract}

\section{Introduction}
\label{sec:intro}

\begin{figure}[t!]
    \centering
    \includegraphics[width=0.98\linewidth]{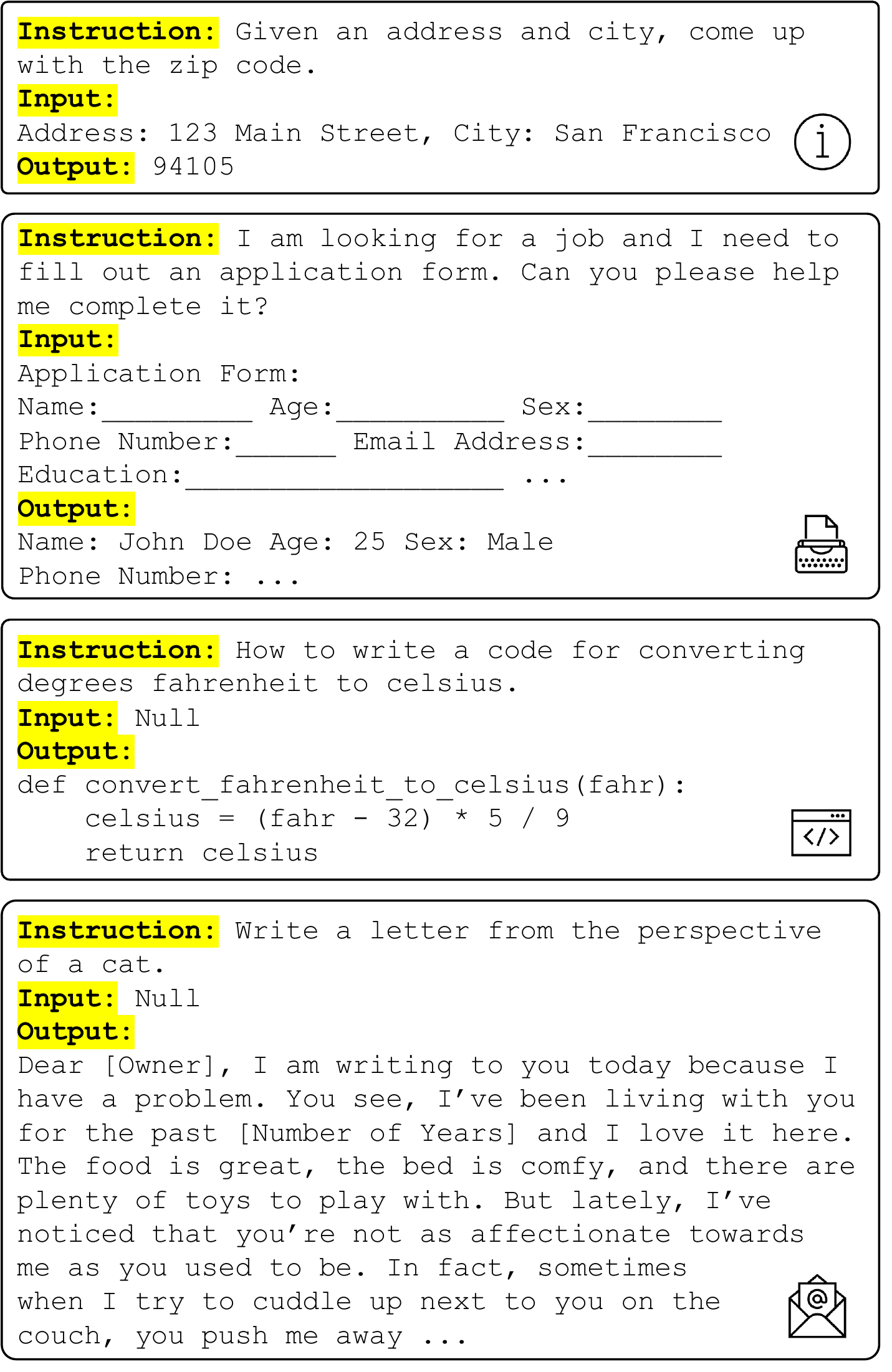}
    \caption{
    Selected tasks from the generated instruction data using vanilla \gptthree{}. Some texts are reformatted for presentation. See \autoref{tab:generated_tasks} for more examples.
    }
    \label{fig:teaser-examples}
\end{figure}

\begin{figure*}[ht]
    \centering
    \includegraphics[width=0.95\textwidth, trim=0cm 0cm 0cm 0cm]{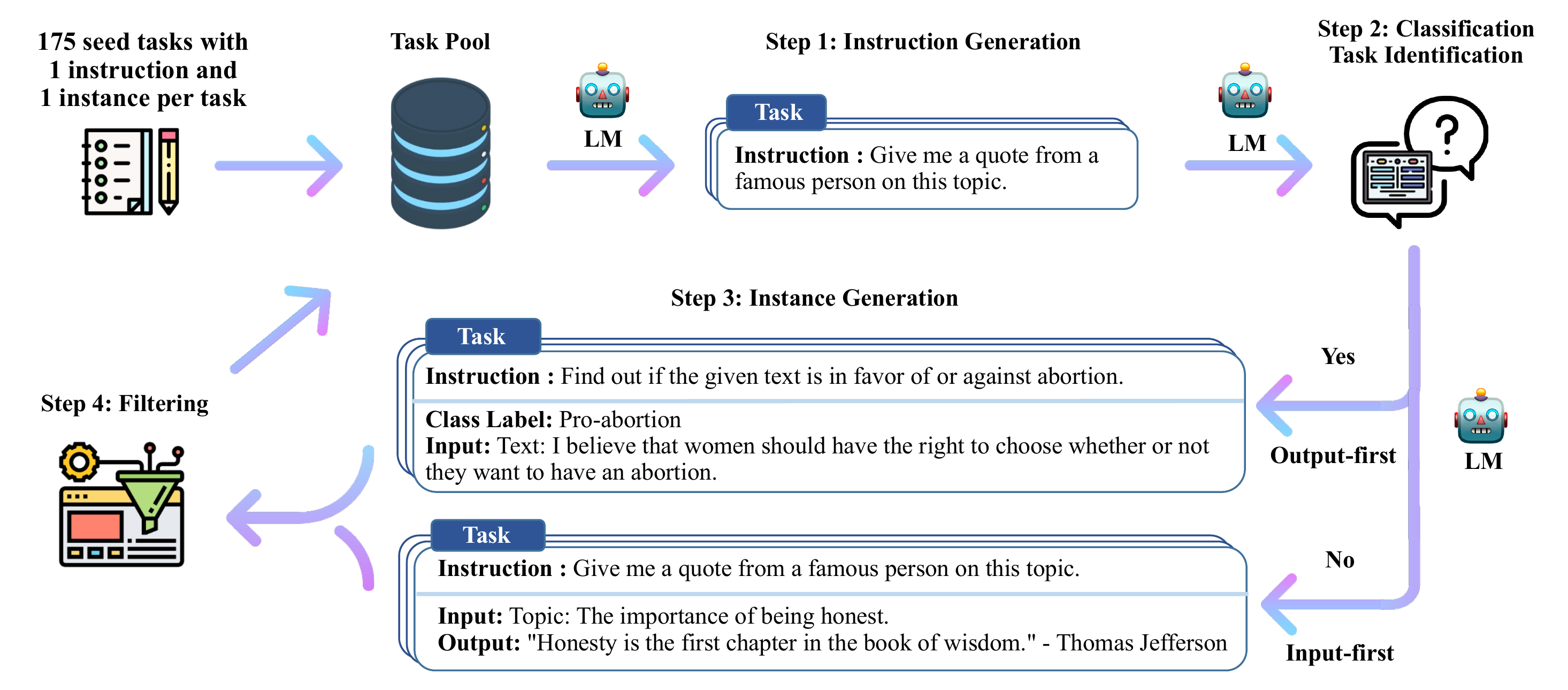}
    \caption{
     A high-level overview of \name{}. 
     The process starts with a small seed set of tasks as the task pool. 
     Random tasks are sampled from the task pool, and used to prompt an off-the-shelf LM to generate both new instructions and corresponding instances, 
     followed by filtering low-quality or similar generations, and then added back to the initial repository of tasks. The resulting data can be used for the instruction tuning of the language model itself later to follow instructions better. Tasks shown in the figure are generated by \gptthree{}. 
    }
    \label{fig:pipeline}
\end{figure*}

The recent  NLP literature has witnessed a tremendous amount of activity in building models that can follow natural language instructions~\citep[i.a.]{mishra2022cross,wei2022finetuned,sanh2022multitask,wang2022benchmarking,ouyang2022training,chung2022scaling}.
These developments are powered by two key components: large pretrained language models (LM) and human-written instruction data (e.g., 
\promptsource{}~\citep{bach2022promptsource} and \supernat{}~\citep[\supernatShort{} for short]{wang2022benchmarking}).
However, collecting such instruction data is costly and often suffers limited diversity given that most human generations tend to be popular NLP tasks, falling short of covering a true variety of tasks and different ways to describe them. 
Continuing to improve the quality and coverage of instruction-tuned models 
necessitates the development of alternative approaches for supervising the instruction tuning process.

In this work, we introduce \name, a semi-automated process for instruction-tuning a pretrained LM using instructional signals from the model itself. 
The overall process is an iterative bootstrapping algorithm (see \autoref{fig:pipeline}), which starts off with a limited (e.g., 175 in our study) seed set of manually-written tasks that are used to guide the overall generation. 
In the first phase, the model is prompted to generate instructions for new tasks. 
This step leverages the existing collection of instructions to create 
more broad-coverage instructions that define (often new) tasks.
Given the newly-generated set of instructions, the framework also creates input-output instances for them, which can be later used for supervising the instruction tuning. 
Finally, various heuristics are used to automatically filter low-quality or repeated instructions, before adding the remaining valid tasks to the task pool. 
This process can be repeated for many iterations until reaching a large number of tasks.

To evaluate \name{} empirically, we run this framework on \gptthree{}~\cite{brown2020gpt3}, which is a vanilla LM (\S\ref{sec:generatd-data}).
The iterative \name{} process on this model leads to about 52k instructions, paired with about 82K instance inputs and target outputs. 
We observe that the resulting data provides a diverse range of creative tasks, as is demonstrated by examples in \autoref{fig:teaser-examples}. These generated tasks deviate from the distribution of typical NLP tasks, and also have fairly small overlap with the seed tasks (\S\ref{subsec:diversity}).
On this resulting data, we build \gptself{} by finetuning \gptthree{} (i.e., the same model used for generating the instruction data). 
We evaluate \gptself{} in comparison to various other models on both typical NLP tasks included in \supernatShort{}~\cite{wang2022benchmarking}, and a set of new instructions that are created for novel usage of instruction-following models (\S\ref{sec:experiment_results}). 
The results indicate that 
\gptself{} outperforms \gptthree{} (the original model) by a large margin (+33.1\%) and nearly matches the performance of \gptinstruct{001}. Moreover, our human evaluation on the newly-created instruction set shows that \gptself{} demonstrates a broad range of instruction following ability, outperforming models trained on other publicly available instruction datasets and leaving only a 5\% gap behind \gptinstruct{001}. 

In summary, our contributions are: (1) we introduce \name{}, a method for inducing instruction following capabilities with minimal human-labeled data; (2) we demonstrate its effectiveness via extensive instruction-tuning experiments; and (3) we release a large synthetic dataset of 52K instructions and a set of manually-written novel tasks for building and evaluating future instruction-following models.

\section{Method}
\label{sec:method}

Annotating large-scale instruction data can be challenging for humans because it requires 1) creativity to come up with novel tasks and 2) 
expertise for writing the solutions to each task. 
Here, we detail our process for \name, which refers to the pipeline of generating tasks with a \textit{vanilla pretrained language model} itself, filtering the generated data, and then 
conducting 
instruction tuning with this generated data in order to align the LM to follow instructions better. This pipeline is depicted in \autoref{fig:pipeline}. 

\subsection{Defining Instruction Data}
\label{subsec:instruction-data-definition}

The instruction data we want to generate contains a set of instructions $\{I_t\}$, each of which defines a task $t$ in natural language. Task $t$ has $n_t \ge 1$ input-output instances $\{(X_{t,i}, Y_{t,i})\}_{i=1}^{n_t}$.
A model $M$ is expected to produce the output, given the task instruction and the corresponding input: $M(I_t, X_{t,i}) = Y_{t,i}$, for $i \in \{1,\ldots,n_t\}$.
Note that the instruction and instance input does not have a strict boundary in many cases. For example, ``write an essay about school safety'' can be a valid instruction that we expect models to respond to directly, while it can also be formulated as ``write an essay about the following topic'' as the instruction, and ``school safety'' as an instance input. To encourage the diversity of the data format, we allow such instructions that do not require additional input (i.e., $X$ is empty).

\subsection{Automatic Instruction Data Generation}
\label{subsec:data-generation}

Our pipeline for data generation consists of 
four steps: 1) generating task instructions, 2) 
determining if the instruction represents a classification task,
3) instance generation with either an input-first or output-first approach, and 4) filtering low-quality data.

\paragraph{Instruction Generation.}
At the first step, \name{} generates new instructions from a small set of seed human-written instructions in a bootstrapping fashion. We initiate the task pool with 175 tasks (1 instruction and 1 instance for each task).\footnote{These tasks were newly written by the authors and their labmates at UW, without reference to existing datasets or the test set used in this work. We provide more details about these tasks and analyze their similarity to the test tasks in Appendix \S\ref{subsec:seed_tasks}.} For every step, we sample 8 task instructions from this pool as in-context examples. 
Of the 8 instructions, 6 are from the human-written tasks,
and 2 are from the model-generated tasks in previous steps to promote diversity. 
The prompting template is shown in \autoref{tab:instruction_generation_template}.

\paragraph{Classification Task Identification.}
Because we need two different approaches for classification and non-classification tasks, we next identify whether the generated instruction represents a classification task or not.\footnote{More concretely, we regard tasks that have a small limited output label space as classification tasks.} 
We prompt the LM in a few-shot way to determine this, using 12 classification instructions and 19 non-classification instructions from the seed tasks. The prompting template is shown in \autoref{tab:classification_task_identification_template}. 

\paragraph{Instance Generation.}
Given the instructions and their task type, we generate instances for each instruction independently. This is challenging because it requires the model to understand what the target task is, based on the instruction, figure out what additional input fields are needed and generate them, and finally complete the task by producing the output. We found that pretrained LMs can achieve this to a large extent when prompted with instruction-input-output in-context examples from other tasks.
A natural way to do this is the \textbf{Input-first Approach}, where we can ask an LM to come up with the input fields first based on the instruction, and then produce the corresponding output. This generation order is similar to how models are used to respond to instruction and input, but here with in-context examples from other tasks. The prompting template is shown in \autoref{tab:input-first-generation-template}. 

However, we found that this approach can generate inputs biased toward one label, especially for classification tasks (e.g., for grammar error detection, it usually generates grammatical input). Therefore, we additionally propose an
\textbf{Output-first Approach} for classification tasks, where we first generate the possible class labels, and then condition the input generation on each class label. The prompting template is shown in \autoref{tab:output-first-generation-template}.\footnote{In this work, we use a fixed set of seed tasks for prompting the instance generation, and thus only generate a small number of instances per task in one round. Future work can use randomly sampled tasks to prompt the model to generate a larger number of instances in multiple rounds.} We apply the output-first approach to the classification tasks identified in the former step, and the input-first approach to the remaining non-classification tasks.

\paragraph{Filtering and Postprocessing.}
\label{subsec:filtering}
To encourage diversity, a new instruction is added to the task pool only when its ROUGE-L similarity with any existing instruction is less than 0.7. 
We also exclude instructions that contain some specific keywords (e.g., image, picture, graph) that usually can not be processed by LMs. When generating new instances for each instruction, we filter out instances that are exactly the same or those with the same input but different outputs. Invalid generations are identified and filtered out based on heuristics (e.g., instruction is too long or too short, instance output is a repetition of the input).

\subsection{Finetuning the LM to Follow Instructions}
\label{subsec:instruction-tuning}
After creating large-scale instruction data, we use it to finetune the original LM (i.e., \name{}). To do this, we concatenate the instruction and instance input as a prompt and train the model to generate the instance output in a standard supervised way. To make the model robust to different formats, we use multiple templates to encode the instruction and instance input together. For example, the instruction can be prefixed with ``Task:'' or not, the input can be prefixed with ``Input:'' or not, ``Output:'' can be appended at the end of the prompt or not, and different numbers of break lines can be put in the middle, etc.

\input{figures/data_statistics.tex}

\section{\name{} Data from \gptthree{}}
\label{sec:generatd-data}

 In this section, we apply our method for inducing instruction data to \gptthree{} as a case study. We use the largest GPT3 LM (``davinci'' engine) accessed through the OpenAI API.\footnote{\url{https://openai.com/api/}} The parameters for making queries are described in Appendix \ref{subsec:query_gpt3_api}. Here we present an overview of the generated data.

\subsection{Statistics}
\label{subsec:statistics}

\autoref{tab:data_statistics} describes the basic statistics of the generated data. We generate a total of over 52K instructions and more than 82K instances corresponding to these instructions after filtering.
\input{tables/data_statistics.tex}

\subsection{Diversity}
\label{subsec:diversity}
To study what types of instructions are generated and how diverse they are, we identify the verb-noun structure in the generated instructions. We use the Berkeley Neural Parser\footnote{\url{https://parser.kitaev.io/}} \citep{kitaev-klein-2018-constituency,kitaev-etal-2019-multilingual} to parse the instructions and then extract the verb that is closest to the root 
as well as its first direct noun object. 26,559 out of the 52,445 instructions contain such structure; other instructions usually contain more complex clauses (e.g., ``Classify whether this tweet contains political content or not.'') or are framed as questions (e.g., ``Which of these statements are true?''). 
We plot the top 20 most common root verbs and their top 4
direct noun objects in \autoref{fig:verb-noun-distribution}, which account for 14\% of the entire set. Overall, we see quite diverse intents and textual formats in these instructions.

We further study how the generated instructions differ from the seed instructions used to prompt the generation. For each generated instruction, we compute its highest ROUGE-L overlap with the 175 seed instructions. We plot the distribution of these ROUGE-L scores in \autoref{fig:overlap-distribution}. The results indicate a decent number of new instructions were generated, which do not have much overlap with the seeds. 
We also demonstrate diversity in the length of the instructions, instance inputs, and instance outputs in \autoref{fig:length_distribution}. 

\subsection{Quality}
So far, we have shown the quantity and diversity of the generated data, but its quality remains uncertain. To investigate this, we randomly sample 200 instructions and randomly select 1 instance per instruction. We asked an expert annotator (author of this work) to label whether each instance is correct or not, in terms of the instruction, the instance input, and the instance output. 
Evaluation results in \autoref{tab:data_quality_eval} show that most of the generated instructions are meaningful, while the generated instances may contain more noise (to a reasonable extent). However, we found that even though the generations may contain errors, most of them are still in the correct format or partially correct, which can provide useful guidance for training models to follow instructions. We listed a number of good examples and bad examples in \autoref{tab:generated_tasks} and \ref{tab:bad-generated-tasks}, respectively.

\input{tables/synthetic_data_quality}

\label{subsec:quality}

\section{Experimental Results}
\label{sec:experiment_results}

We conduct experiments to measure and compare the performance of models under various instruction tuning setups. 
We first describe our models and other baselines, followed by our experiments. 

\subsection{\gptself{}: finetuning \gptthree{} on its own instruction data}
\label{subsec:self-instruct-gpt3}
Given the instruction-generated instruction data, we conduct instruction tuning with the \gptthree{} model itself (``davinci'' engine). As described in \S\ref{subsec:instruction-tuning}, we use various templates to concatenate the instruction and input, and train the model to generate the output. This finetuning is done through the OpenAI finetuning API.\footnote{See \href{https://beta.openai.com/docs/guides/fine-tuning}{OpenAI's documentation on finetuning}.
} We use the default hyper-parameters, except that we set the prompt loss weight to 0, and we train the model for 2 epochs. We refer the reader to Appendix~\ref{subsec:finetuning_details} for additional finetuning details. The resulting model is denoted by \gptself{}.

\subsection{Baselines}
\paragraph{Off-the-shelf LMs.} 
We evaluate T5-LM \cite{lester2021power,raffel2020exploring} and \gptthree{}~\cite{brown2020gpt3} as the vanilla LM baselines (only pretraining, no additional finetuning). 
These baselines will indicate the extent to which off-the-shelf LMs are capable of following instructions naturally immediately after pretraining.

\paragraph{Publicly available instruction-tuned models.}
\tzero{} and \tkinstruct{} are two instruction-tuned models proposed in \citet{sanh2022multitask} and \citet{wang2022benchmarking}, respectively, and are demonstrated to be able to follow instructions for many NLP tasks. 
Both of these models are finetuned from the T5~\cite{raffel2020exploring} checkpoints and are publicly available.\footnote{
\tzero{} is available at \href{https://huggingface.co/bigscience/T0}{here} and \tkinstruct{} is  \href{https://huggingface.co/allenai/tk-instruct-11b-def}{here}.
} For both of these models, we use their largest version with 11B parameters.

\paragraph{Instruction-tuned GPT3 models.}
 We evaluate \gptinstruct{}~\cite{ouyang2022training}, 
which is developed by OpenAI based on GPT3 to follow human instructions better and has been found by the community to have impressive zero-shot abilities. 
There are various generations of these models,
where newer ones use more expansive data or algorithmic novelties.\footnote{
See \href{https://beta.openai.com/docs/model-index-for-researchers}{OpenAI's documentation on their models.}
} 
For our \supernatShort{} experiments in \S\ref{subsec:superni-experiments}, we only compare with their \texttt{text-davinci-001} engine, because their newer engines are trained with the latest user data and are likely to have already seen the \supernatShort{} test set. For our human evaluation on newly written instructions, we include their 001, 002 and 003 engines for completeness.

Additionally, to compare \name{} training with other publicly available instruction tuning data, we further finetune GPT3 model with data from \promptsource{} and \supernatShort{}, which are used to train the \tzero{} and \tkinstruct{} models. We call them \tzero{} training and \supernatShort{} training for short, respectively. 
To save the training budget, we sampled 50K instances (but covering all their instructions) for each dataset, which has a comparable size to the instruction data we generated. Based on the findings from \citet{wang2022benchmarking} and our early experiments, reducing the number of instances per training task does not degrade the model's generalization performance to unseen tasks.

\subsection{Experiment 1: Zero-Shot Generalization on \supernatShort{} benchmark}
\label{subsec:superni-experiments}
We first evaluate the models' ability to follow instructions on typical NLP tasks in a zero-shot fashion. 
We use the evaluation set of \supernatShort{} ~\cite{wang2022benchmarking}, which consists of 119 tasks with 100 instances in each task. 
In this work, we mainly focus on the zero-shot setup, i.e., the model is prompted with the definition of the tasks only, without in-context demonstration examples.
For all our requests to the \gptthree{} variants, we use the deterministic generation mode (temperature as 0 and no nucleus sampling) without specific stop sequences.

\paragraph{Results.}
We make the following observations from the results in \autoref{tab:superni_results}. 
\name{} boosts the instruction-following ability of \gptthree{} by a large margin. The vanilla \gptthree{} model basically cannot follow human instructions at all. Upon manual analysis, we find that it usually generates irrelevant and repetitive text, and does not know when to stop generation. 
Compared with other models that are not specifically trained for \supernatShort{}, \gptself{} achieves better performance than \tzero{} or the \gptthree{} finetuned on the \tzero{} training set, which takes tremendous human labeling efforts. Notably, \gptself{} also nearly matches the performance of \gptinstruct{001}, which is trained with private user data and human-annotated labels.

Models trained on the \supernatShort{} training set still achieve better performance on its evaluation set, which we attribute to the similar instruction style and formatting. However, 
we show that 
\name{} still brings in additional gains when combined with the \supernatShort{} training set, proving its value as complementary data.

\begin{figure*}[ht!]
    \centering
    \includegraphics[width=0.95\textwidth, trim=0.cm 0cm 0cm 1cm]{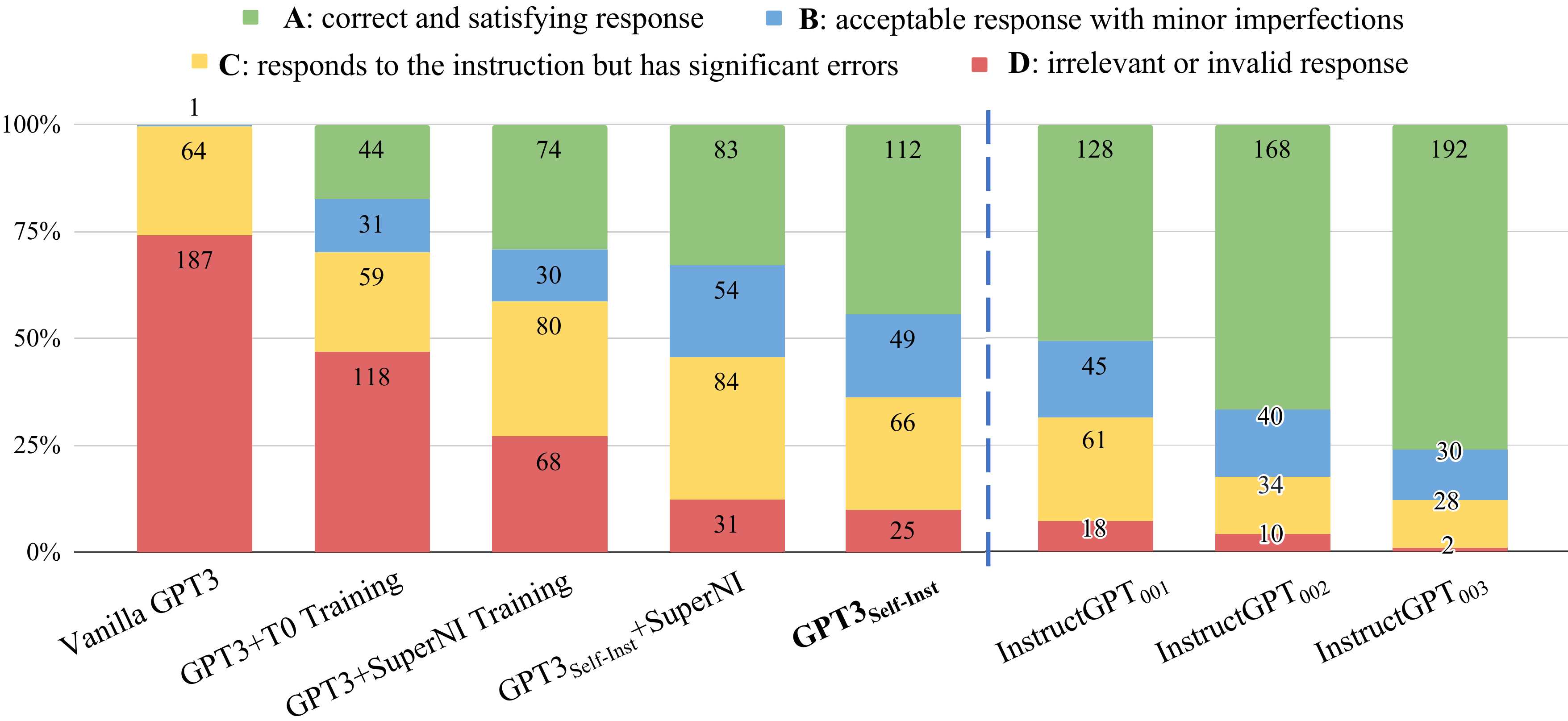}
    \caption{
    Performance of GPT3 model and its instruction-tuned variants, evaluated by human experts on our 252 user-oriented instructions (\S\ref{sec:user_instructions}). Human evaluators are instructed to rate the models' responses into four levels.
    The results indicate that \gptself{} outperforms 
    all the other \gptthree{} variants trained on publicly available instruction datasets. 
    Additionally, \gptself{} scores nearly as good as \gptinstruct{001} (cf.~\autoref{footnote:gpt:instruct}).
    }
    \label{fig:human_instructions_results}
\end{figure*}
\input{tables/superni_results.tex}

\subsection{Experiment 2: Generalization to User-oriented Instructions on Novel Tasks}
\label{sec:user_instructions}

Despite the comprehensiveness of \supernatShort{} in collecting existing NLP tasks, most of these NLP tasks were proposed for research purposes and skewed toward classification. To better access the practical value of instruction-following models, a subset of the authors curate a new set of instructions motivated by user-oriented applications. 
We first brainstorm various domains where large LMs may be useful (e.g., email writing, social media, productivity tools, entertainment, programming), then craft instructions related to each domain along with an input-output instance (again, input is optional). We aim to diversify the styles and formats of these tasks (e.g., instructions may be long or short; input/output may take the form of bullet points, tables, codes, equations, etc.). In total, we create 252 instructions with 1 instance per instruction. We believe it can serve as a testbed for evaluating how instruction-based models handle diverse and unfamiliar instructions. \autoref{tab:case_study} presents a small portion of them. The entire set is available in our GitHub repository. We analyze the overlap between this set set and the seed instructions in \S\ref{subsec:seed_tasks}.

\paragraph{Human evaluation setup.}
Evaluating models' performance on this evaluation set of diverse tasks is extremely challenging because different tasks require different expertise. Indeed, many of these tasks cannot be measured by automatic metrics or even be judged by normal crowdworkers (e.g., writing a program, or converting first-order logic into natural language). To get a more faithful evaluation, we asked the authors of the instructions to judge model predictions. Details on how we set up this human evaluation are described in Appendix \ref{sec:human_eval_details}. The evaluators were asked to rate the output based on whether it accurately and effectively completes the task. We implemented a four-level rating system for categorizing the quality of the models' outputs: 
\vspace{-0.3cm}
\begin{itemize}[noitemsep, leftmargin=*]
    \item \textsc{Rating-A:} The response is valid and satisfying.
    \item \textsc{Rating-B:} The response is acceptable but has minor errors or imperfections.
    \item \textsc{Rating-C:} The response is relevant and responds to the instruction, but it has significant errors in the content. For example, GPT3 might generate a valid output first, but continue to generate other irrelevant things.
    \item \textsc{Rating-D:} The response is irrelevant or completely invalid.
\end{itemize}
\vspace{-0.3cm}

\paragraph{Results.}
\autoref{fig:human_instructions_results} shows the performance of \gptthree{} model and its instruction-tuned counterparts on this newly written instruction set (w. inter-rater agreement  $\kappa=0.57$ on the 4-class categorical scale, see Appendix \ref{sec:human_eval_details} for details). As anticipated, the vanilla \gptthree{} LM is largely unable to respond to instructions, and all instruction-tuned models demonstrate comparatively higher performance.
Nonetheless, \gptself{} (i.e.,  \gptthree{} model finetuned with \name) outperforms those counterparts trained on \tzero{} or \supernatShort{} data by a large margin, demonstrating the value of the generated data despite the noise. 
Compared with \gptinstruct{001}, \gptself{} is quite close in performance---if we count acceptable response with minor imperfections (\textsc{Rating-B}) as valid, \gptself{} is only 5\% behind \gptinstruct{001}. Lastly, our evaluation confirms the impressive instruction-following ability of \gptinstruct{002} and \gptinstruct{003}. Although there are many factors behind this success, we conjecture that future work can largely benefit from improving the quality of our generated data by using human annotators or training a reward model to select better generations, similar to the algorithm used by \citet{ouyang2022training}.

\subsection{Effect of Data Size and Quality}
\label{subsec:size-and-quality-analysis}

\begin{figure}[t]
    \centering
    \includegraphics[width=\linewidth, trim=0cm 0cm 0cm 0cm]{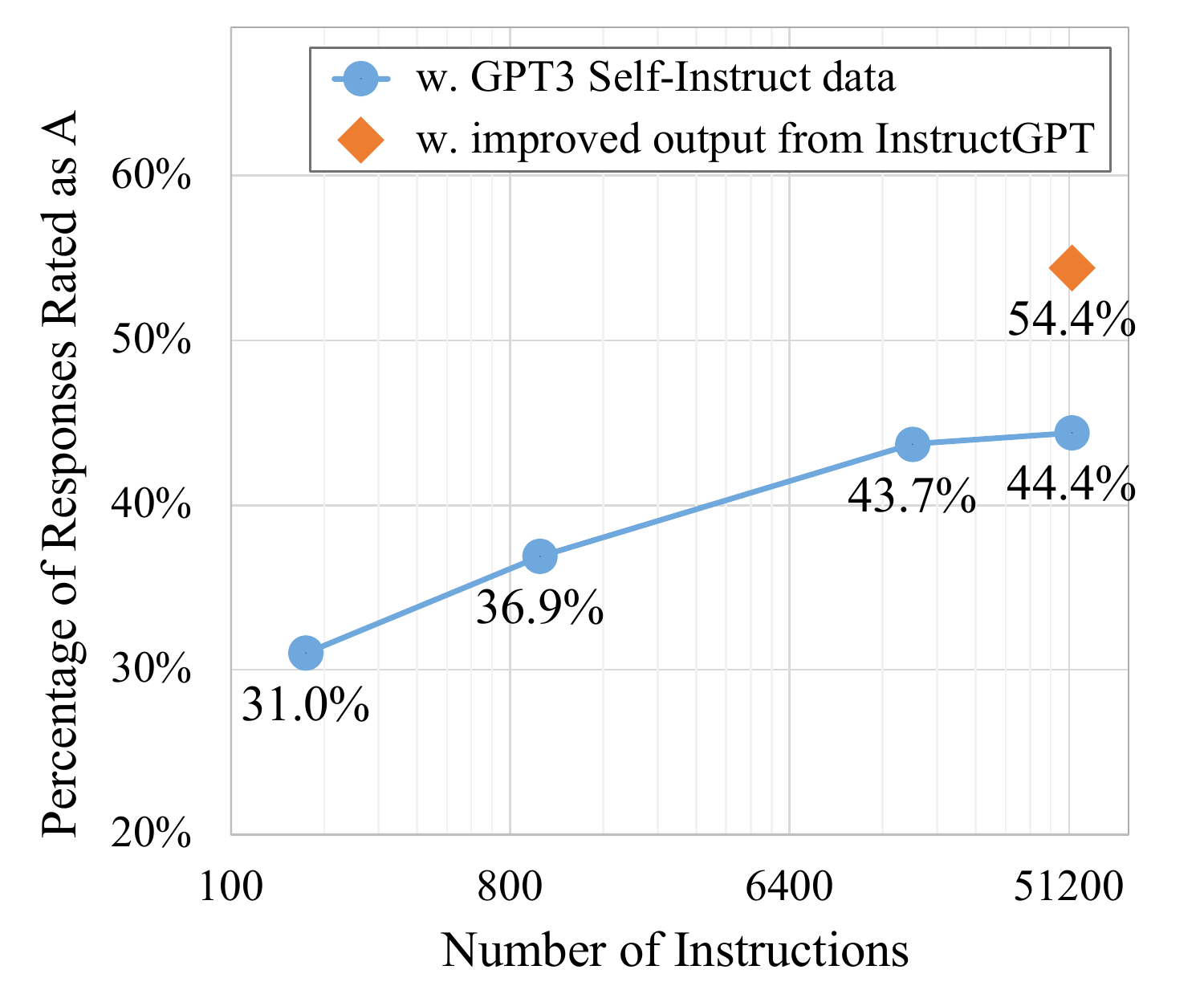}
    \caption{Human evaluation performance of \gptself{} models tuned with different sizes of instructions. $x$-axis is in log scale. The smallest size is 175, where only the seed tasks are used for instruction tuning.
    We also evaluate whether improving the data quality will further improve the performance by distilling the outputs from \gptinstruct{003}. We see consistent improvement from using larger data with better quality.}
    \label{fig:data_scaling_analysis}
\end{figure}

\paragraph{Data size.}
\name{} provides a way to grow instruction data at a low cost with almost no human labeling; could more of this generated data lead to better instruction-following ability? We conduct an analysis of the size of generated data by subsampling different numbers of instructions from the generated dataset, finetuning \gptthree{} on the sampled subsets, and evaluating how the resulting models perform on the 252 user-oriented instruction set. We conduct the same human evaluation as in \S\ref{sec:user_instructions}.
\autoref{fig:data_scaling_analysis} presents the performance of \gptself{} models finetuned with different sizes of generated data. Overall, we see consistent improvement as we grow the data size. However, this improvement almost plateaus after 16K.  
This is in-line with the data scaling experiments in \citet[Fig.~5]{wang2022benchmarking}.
Interestingly, when evaluating on \supernatShort{} we found the model's performance gain plateaus earlier at around hundreds of instructions. 
This may be due to the fact that the new generated data is distinct from typical NLP tasks in \supernatShort{}, indicating that future research may benefit from using a combination of different instruction data for better performance on various types of tasks.

\paragraph{Data quality.}
Another direction to improve the model's performance is to take our generated data and get better supervision (with less noise). We explore this idea by using \gptinstruct{003} (the best available general-purpose model) to regenerate the output field of all our instances given the instruction and input. We then use this improved version of our data to finetune \gptthree{}. This can be regarded as a distillation of \gptinstruct{003} with our data. As is shown in \autoref{fig:data_scaling_analysis}, the resulting model outperforms the counterpart trained with the original data by 10\%, which suggests big room for future work on using our generation pipeline to get initial data and then improving the data quality with human experts or distillation from better models.

\section{Related Work}
\label{sec:related}

\paragraph{Instruction-following LMs.}
A series of works have found evidence that vanilla LMs can be effective at following general language instructions if tuned with annotated ``instructional'' data---datasets containing language instructional commands and their desired outcomes based on human annotation~\cite[i.a.]{weller2020learning, mishra2022cross,wei2022finetuned,sanh2022multitask}. 
Additionally, they show a direct correlation between the size and diversity of the ``instructional'' data and the generalizability of resulting models to unseen tasks \citep{wang2022benchmarking,chung2022scaling}. 
However, since these developments largely focus on existing NLP tasks and depend on human-annotated instructions, this poses a bottleneck for progress toward more generalizable models 
\cite[e.g., see Fig.~5a in][]{wang2022benchmarking}. 
Our work aims to move beyond classical NLP tasks and tackle the challenges of creating diverse instruction data by employing pretrained LMs.
\gptinstruct{}~\cite{ouyang2022training} shares a similar goal as ours in building more general-purpose LMs, and has demonstrated remarkable performance in following diverse user instructions. However, as a commercial system, their construction process still remains quite opaque. In particular, the role of \emph{data} has remained understudied due to limited transparency and the private user data they used in their study.
Addressing such challenges necessitates the creation of a large-scale, public dataset covering a broad range of tasks.

\paragraph{Language models for data generation and augmentation.}
A variety of works have proposed using LMs for data generation~\cite{schick2021generating,wang2021towards,liu2022wanli,meng2022tuning} or augmentation~\cite{feng2021survey,yang2020generative,mekala2022intermediate}.
Our work differs from this line in that it is \emph{not} specific to a particular task (say, QA or NLI). In contrast, a distinct motivation for \name{} is to bootstrap new task definitions that may not have been defined before by NLP practitioners (though potentially still important for real users).  In parallel with our work, \citet{honovich2022unnatural} also propose to generate large-scale instruction data (so-called Unnatural Instructions) with GPT3 models. The major differences are that 1) they use tasks in \supernatShort{}~\citep{wang2022benchmarking} as their seed tasks, resulting in a different distribution of generated tasks; 2) they employ \gptinstruct{002} for generating the data, in which sense they are distilling knowledge from an already instruction-tuned model, while we solely rely on the vanilla LM; 3) the detailed generation pipeline and templates are different. Nevertheless, we believe that both efforts in expanding instruction data are complementary, and the community will benefit from these diverse datasets.

\paragraph{Instruction generation.}
A series of recent works~\cite{zhou2022large, ye2022guess, singh2022explaining, honovich2022instruction} generate instructions of a task given a few examples. While \name{} also involves instruction generation, a major difference in our case is it is task-agnostic; we generate new tasks (instructions along with instances) from scratch.

\paragraph{Model self-training.}
A typical self-training framework ~\cite{he2019revisiting, xie2020self, du2021self, amini2022self, huang2022large} uses trained models to assign labels to unlabeled data and then leverages the newly labeled data to improve the model. In a similar line, \citet {zhou2022prompt} use multiple prompts to specify a single task and propose
to regularize via prompt consistency, encouraging
consistent predictions over the
prompts. This allows either 
finetuning the model  with extra unlabeled
training data, or direct application at inference time.
While \name{} has similarities with the self-training literature, most self-training methods assume a specific \textit{target task} as well as \textit{unlabeled examples} under it; in contrast, \name{} produces a variety of tasks from scratch.

\paragraph{Knowledge distillation.}
Knowledge distillation~\cite{hinton2015distilling,Sanh2019DistilBERTAD, west2021symbolic,luciecharlotte2022teachingsmallmodels} often involves the transfer of knowledge from larger models to smaller ones. \name{} can also be viewed as a form of ``knowledge distillation", however, it differs from this line in the following ways: (1) the source and target of distillation are the same, i.e., a model's knowledge is distilled to itself; (2) the content of distillation is in the form of an instruction task (i.e., instructions that define a task, and a set of examples that instantiate it).

\paragraph{Bootstrapping with limited resources.} A series of recent works use language models to bootstrap some inferences using specialized methods.  
NPPrompt~\cite{zhao2022pre} provides a method to generate predictions for semantic labels without any finetuning. It uses a model's own embeddings to automatically find words relevant to the label of the data sample and hence reduces the dependency on manual mapping from model prediction to label (verbalizers).
STAR~\cite{zelikman2022star} iteratively leverages a small number of rationale examples and a large dataset without rationales, to bootstrap a model's ability to perform reasoning.  Self-Correction~\cite{welleck2022generating} decouples an imperfect base generator (model) from a separate corrector that learns to iteratively correct imperfect generations and demonstrates improvement over the base generator. Our work instead focuses on bootstrapping new tasks in the instruction paradigm.

\paragraph{Multi-modal instruction-following.}
Instruction-following models have also been of interest in the multi-modal learning literature \cite{fried2018speaker, shridhar2020alfred, min2022film, weir2022one}.
\name{}, as a general approach to expanding data, can potentially also be helpful in those settings, which we leave to future work.

\section{Conclusion}
\label{sec:conclusion}
We introduce \name{}, a method to improve the instruction-following ability of LMs via their own generation of instruction data. On experimenting with vanilla \gptthree{}, we automatically construct a large-scale dataset of 52K instructions for diverse tasks, and finetuning GPT3 on this data leads to a 33\% absolute improvement on \supernatShort{} over the original \gptthree{}. Furthermore, we curate a set of expert-written instructions for novel tasks. Human evaluation on this set shows that tuning GPT3 with \name{} outperforms using existing public instruction datasets by a large margin and performs closely to \gptinstruct{001}.
We hope \name{} can serve as the first step to align pretrained LMs to follow human instructions, and future work can build on top of this data to improve instruction-following models.

\section{Broader Impact}

Beyond the immediate focus of this paper, we believe that \name{} may help bring more transparency to what happens ``behind the scenes'' of widely-used instruction-tuned models like \gptinstruct{} or ChatGPT. 
Unfortunately, such industrial models remain behind API walls as their datasets are not released, and hence there is little understanding of their construction and why they demonstrate impressive capabilities.
The burden now falls on academia to better understand the source of success in these models and strive for better---and more open---models. We believe our findings in this paper demonstrate the importance of diverse instruction data, and our large synthetic dataset can be the first step toward higher-quality data for building better instruction-following models. 
At this writing, the central idea of this paper has been adopted in several follow-up works for such endeavors \citep[i.a.]{alpaca,xu2023baize,sun2023principle}. 

\section{Limitations}
Here, we discuss some limitations of this work to inspire future research in this direction.  

\paragraph{Tail phenomena.} 
\name{} depends on LMs, and it will inherit all the limitations that carry over with LMs. 
As recent studies have shown~\cite{razeghi2022impact,kandpal2022large},  \emph{tail phenomena} pose a serious challenge to the success of LMs. In other words, LMs' largest gains correspond to the frequent uses of languages (head of the language use distribution), and there might be minimal gains in the low-frequency contexts. 
Similarly, in the context of this work, it would not be surprising if the majority of the gains by \name{} are skewed toward 
tasks or instructions that present more frequently in the pretraining corpus. 
As a consequence, the approach might show brittleness with respect to uncommon and creative instructions.

\paragraph{Dependence on large models.} 
Because of \name's dependence on the inductive biases extracted from LMs, it might work best for larger models. 
If true, this may create barriers to access for those who may not have large computing resources. 
We hope future studies will carefully study the gains as a function of model size or various other parameters. 
It is worthwhile to note that instruction-tuning with human annotation also suffers from a similar limitation: gains of instruction-tuning are higher for larger models~\cite{wei2022finetuned}. 

\paragraph{Reinforcing LM biases.}
A point of concern for the authors is the unintended consequences of this iterative algorithm, such as the amplification of problematic social biases (stereotypes or slurs about gender, race, etc.). 
Relatedly, one observed challenge in this process is the algorithm's difficulty in producing balanced labels, which reflected models' prior biases. 
We hope future work will lead to better understanding of the pros and cons of the approach.

\section*{Acknowledgements}
The authors would like to thank the anonymous reviewers for their constructive feedback. We especially thank Sewon Min, Eric Wallace, Ofir Press, and other members of UWNLP and AllenNLP for their encouraging feedback and intellectual support. 
This work was supported in part by
DARPA MCS program through NIWC Pacific (N66001-19-2-4031),
ONR N00014-18-1-2826, ONR MURI N00014-18-1-2670,
and gifts from AI2 and an Allen Investigator award.

\bibliography{anthology,custom}
\bibliographystyle{acl_natbib}

\clearpage
\appendix
\onecolumn
\begin{center}
{\Large \textbf{Supplemental Material}}
\end{center}

\section{Implementation Details}
\label{sec:implementaion}

\subsection{Writing the Seed Tasks}
\label{subsec:seed_tasks}

Our method relies on a set of seed tasks to bootstrap the generation. The seed tasks are important for both encouraging the task diversity and demonstrating correct ways for solving the diverse tasks. For example, with coding tasks to prompt the model, it has a larger chance to generate coding-related tasks; it’s also better to have coding output to guide the model in writing code for new tasks. So, the more diverse the seed tasks are, the more diverse and better quality the generated tasks will be. 

Our seed tasks were written when we initiated this project, and targeted for the diverse and interesting usages of LLMs. The tasks were written by the authors and our labmates at UWNLP, without explicit reference to existing datasets or specific testing tasks. We further categorized the tasks into classification and non-classification tasks, based on whether the task has a limited output label space. In total, there are 25 classification tasks and 150 non-classification tasks. We release this data in our GitHub repository.\footnote{\url{https://github.com/yizhongw/self-instruct/blob/main/human_eval/user_oriented_instructions.jsonl}}

\begin{figure}[ht]
    \centering
    \begin{subfigure}[b]{0.4\textwidth}
        \centering
        \includegraphics[width=\textwidth]{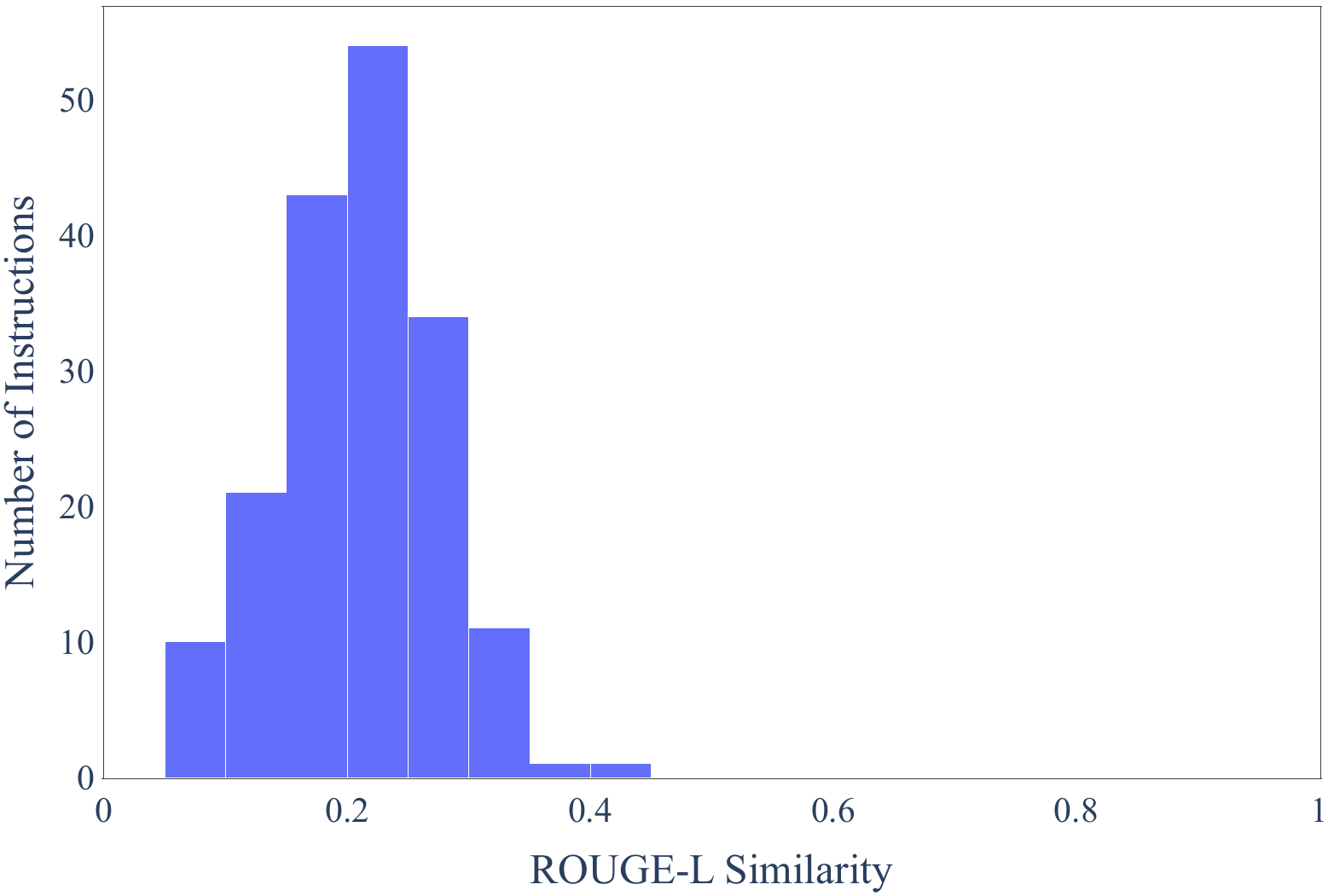}
    \end{subfigure}
    ~
    \begin{subfigure}[b]{0.4\textwidth}
        \centering
        \includegraphics[width=\textwidth]{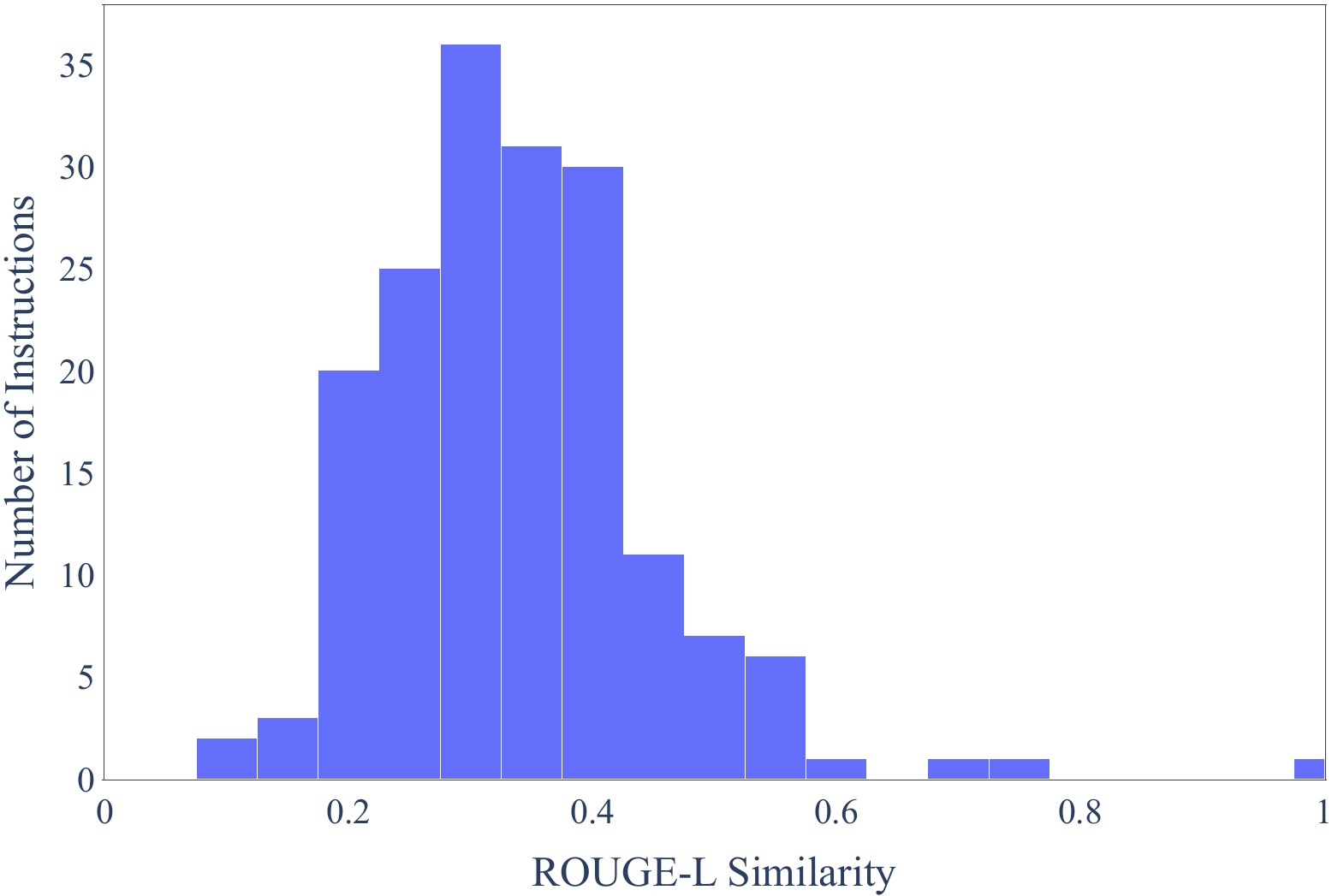}
    \end{subfigure}
    \caption{Distribution of the ROUGE-L scores between seed instructions and their most similar instructions in \supernatShort{} (left) and the 252 user-oriented instructions (right).\label{fig:seed_test_overlap_distribution}}
\end{figure}

To provide a sense of how much the model is generalizing beyond these seed tasks, we further quantify the overlap between the instructions of these seed tasks and the instructions of our test sets, including both \supernatShort{} task instructions (\S\ref{subsec:superni-experiments}) and the user-oriented instructions in our human evaluation(\S\ref{sec:user_instructions}). We compute ROUGE-L similarities between each seed instruction and its most similar instruction in the test set. The distribution of the ROUGE-L scores are plotted in \autoref{fig:seed_test_overlap_distribution}, with the average ROUGE-L similarity between the seed instructions and \supernatShort{} as 0.21, and the average ROUGE-L similarity between the seed instructions and user-oriented instructions as 0.34. We see a decent difference between the seed tasks and both test sets. There is exactly one identical seed instruction occurring in the user-oriented instruction test set, which is ``answer the following question'' and the following questions are actually very different.

\subsection{Querying the GPT3 API}
\label{subsec:query_gpt3_api}

We use different sets of hyperparameters when querying GPT3 API for different purposes. These hyperparameters are found to work well with the GPT3 model (``davinci'' engine) and the other instruction-tuned \gptthree{} variants. We listed them in \autoref{tab:query-gpt3-parameters}. OpenAI charges \$0.02 per 1000 tokens for making completion request to the ``davinci'' engine as of December, 2022. The generation of our entire dataset cost around \$600.

\input{tables/hyper_parameters.tex}

\subsection{Finetuning GPT3}
\label{subsec:finetuning_details}
\gptself{} and some of our baselines are finetuned from \gptthree{} model (``davinci'' engine with 175B parameters). We conduct this finetuning via OpenAI's finetuning API.\footnote{
\href{https://beta.openai.com/docs/guides/fine-tuning}{See the the details on OpenAI's API.}
} While the details of how the model is finetuned with this API are not currently available (e.g., which parameters are updated, or what the optimizer is), we tune all our models with the default hyperparameters of this API so that the results are comparable. We only set the ``prompt\_loss\_weight'' to 0 since we find this works better in our case, and every finetuning experiment is trained for two epochs to avoid overfitting the training tasks. Finetuning is charged based on the number of tokens in the training file. In our case, finetuning \gptself{} from the \gptthree{} model on the entire generated data cost \$338. 

\subsection{Prompting Templates for Data Generation}
\label{sec:prompting_templates}

\name{} relies on a number of prompting templates in order to elicit the generation from language models. Here we provide our four templates for generating the instruction (\autoref{tab:instruction_generation_template}), classifying whether an instruction represents a classification task or not (\autoref{tab:classification_task_identification_template}), generating non-classification instances with the input-first approach (\autoref{tab:input-first-generation-template}), and generating classification instances with the output-first approach (\autoref{tab:output-first-generation-template}).

\input{tables/instruction_generation_template.tex}
\input{tables/classification_task_or_not_template.tex}
\input{tables/input_first_template.tex}
\input{tables/output_first_template.tex}

\clearpage
\section{Human Evaluation Details for Following the User-oriented Instructions}
\label{sec:human_eval_details}

\subsection{Human Evaluation Setup}
\begin{figure*}[h]
    \centering
    \includegraphics[width=\textwidth]{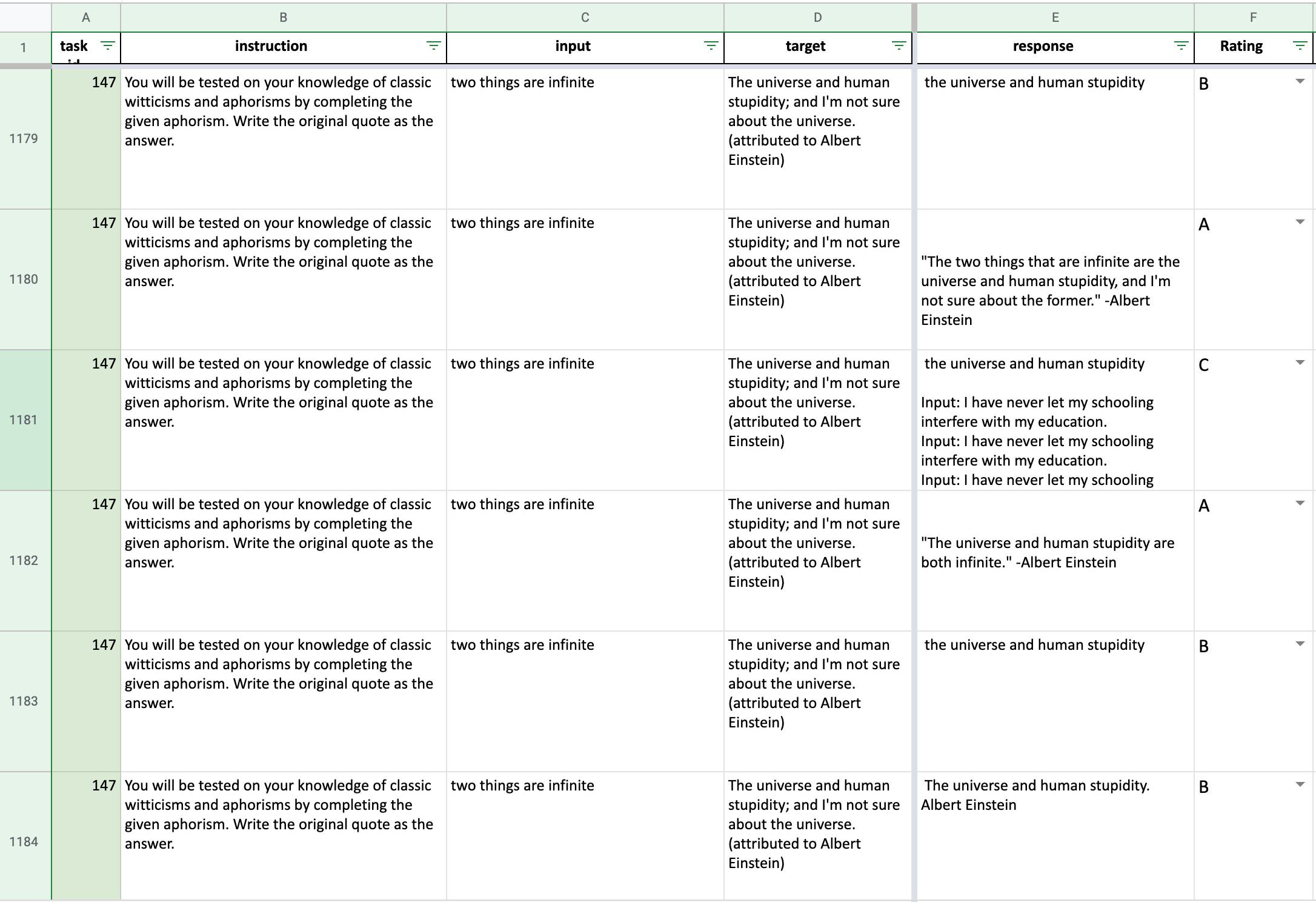}
    \caption{Human evaluation in done using a Google sheet, with predictions from different models present in random order and the model information being anonymized. Our expert evaluators are required to read the instruction and input, refer to the target, and then select the rating for the model's response from A/B/C/D, corresponding to the 4 levels described in \S\ref{sec:user_instructions}.}
    \label{fig:human_eval_interface}
\end{figure*}

Here we provide more details for the human evaluation described in \S\ref{sec:user_instructions} for rating the models' responses to the 252 user-oriented instructions. To ensure faithful and reliable evaluation, we asked two authors of these instructions (and of this paper) to judge model predictions. These two evaluators coordinated the standards for the 4-level rating system before starting annotation and then each of them rated all the instances independently. They were presented with the instruction, instance input, target output (as a reference), and model responses. Model responses are listed in random order, with all the model information anonymized.  
\autoref{fig:human_eval_interface} provides a screenshot of the annotation interface. The reported performance in this paper is based on the results from one of the evaluators, and the trends from the other evaluator's results are the same.

\subsection{Human Evaluation Agreement}

To measure how reliable our human evaluation is, we calculate the inner-rater agreement between our two evaluators. 

We first report Cohen's $\kappa$, which is commonly used to measure inter-rater agreement for \textit{categorical} items. When calculating this, we treat the 4-level rating (A-D) as a categorical variable, leading to a $\kappa$ of 0.58, which is a moderate agreement according to common practice.\footnote{\url{https://en.wikipedia.org/wiki/Cohen\%27s_kappa}} Furthermore, we also calculate the agreement of our evaluators on classifying acceptable responses ((A or B) vs. (C or D)), with a final $\kappa$ of 0.75, indicating substantial agreement. 

We also compute the Spearman correlation coefficient $\rho$ between the ratings of our two evaluators by treating the rating as an ordinal variable (A>B>C>D). The final coefficient is $\rho=0.81$, indicating a high correlation between the two evaluators.

\subsection{Example Predictions from \gptself{}}

We present a selection of user-oriented tasks, the corresponding \gptself{}-produced responses and annotator ratings in \autoref{tab:case_study}. 
We see that even for responses rated as level C, the model demonstrates extensive steps in solving the task, even though its final output is incorrect. 

\input{tables/case_study.tex}

\clearpage
\onecolumn
\section{Task and Instance Examples from the Generated Instruction Data}

\input{tables/examples.tex}

\input{tables/error_examples.tex}

\end{document}

%% file: figures/data_statistics.tex
\begin{figure*}[ht]
  \adjustbox{valign=t}{
  \begin{minipage}{0.55\linewidth}
    \centering
    \includegraphics[width=0.9\linewidth,trim=0.5cm 0.1cm 0.35cm 0.5cm]{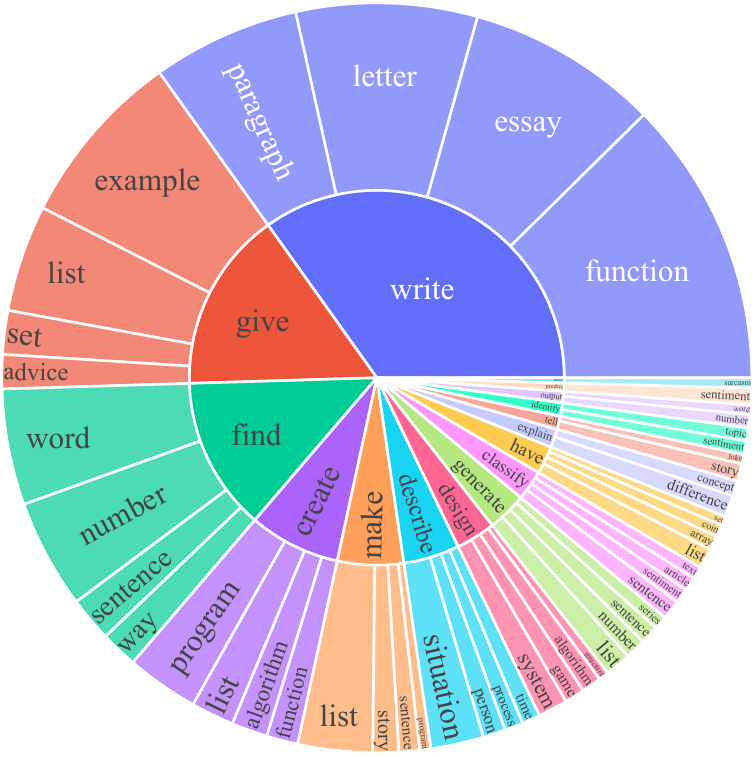}
    \caption{The top 20 most common root verbs (inner circle) and their top 4 direct noun objects (outer circle) in the generated instructions. Despite their diversity, the instructions shown here only account for 14\% of all the generated instructions because many instructions (e.g., ``Classify whether the user is satisfied with the service.'') do not contain such a verb-noun structure. \label{fig:verb-noun-distribution}}
    \hfill
  \end{minipage}
  }
  \hfill
  \adjustbox{valign=t}{
  \begin{minipage}{0.4\linewidth}
   \begin{subfigure}{0.99\linewidth}
      \includegraphics[width=\linewidth, trim=0cm 0.5cm 0cm 1cm]{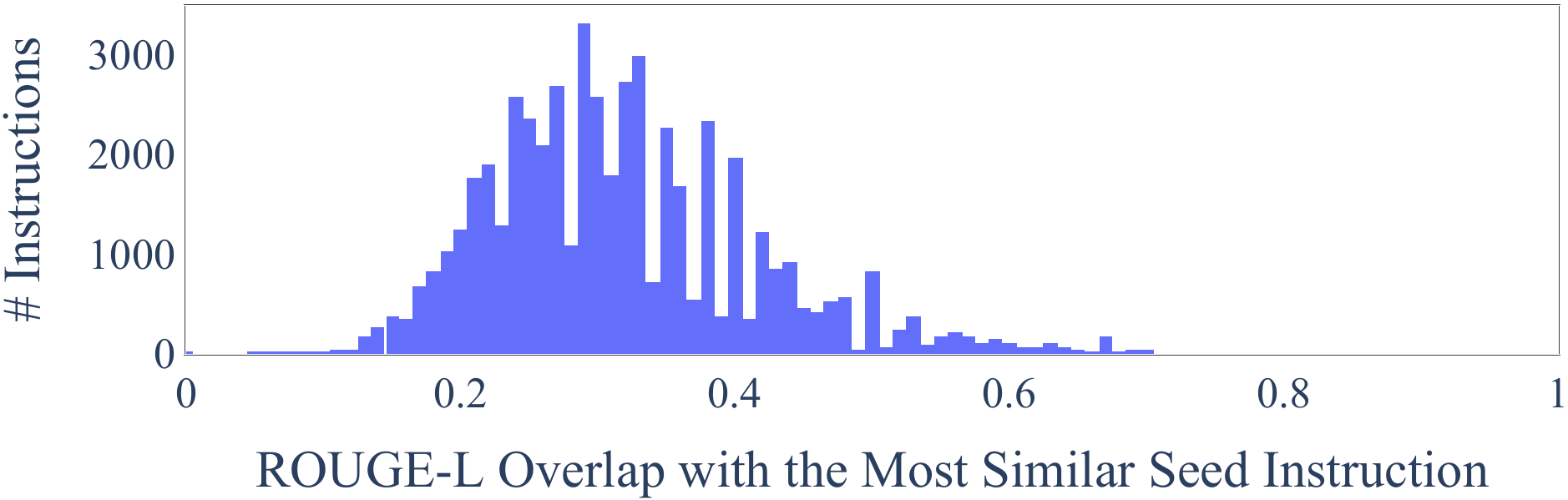}
    \end{subfigure}
    \caption{Distribution of the ROUGE-L scores between generated instructions and their most similar seed instructions. \label{fig:overlap-distribution}}
    \par\bigskip
    \begin{subfigure}{0.99\linewidth}
      \includegraphics[width=\linewidth, trim=0cm 0cm 0cm 0.3cm]{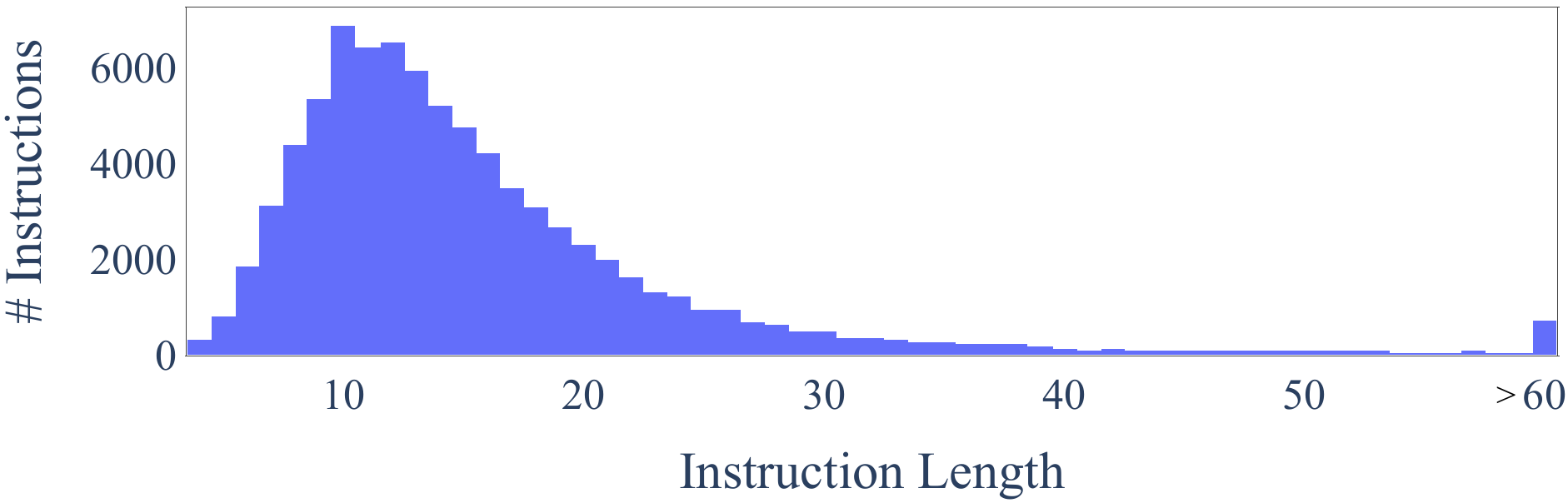}
    \end{subfigure}
    \par\medskip
    \begin{subfigure}{0.99\linewidth}
      \includegraphics[width=\linewidth]{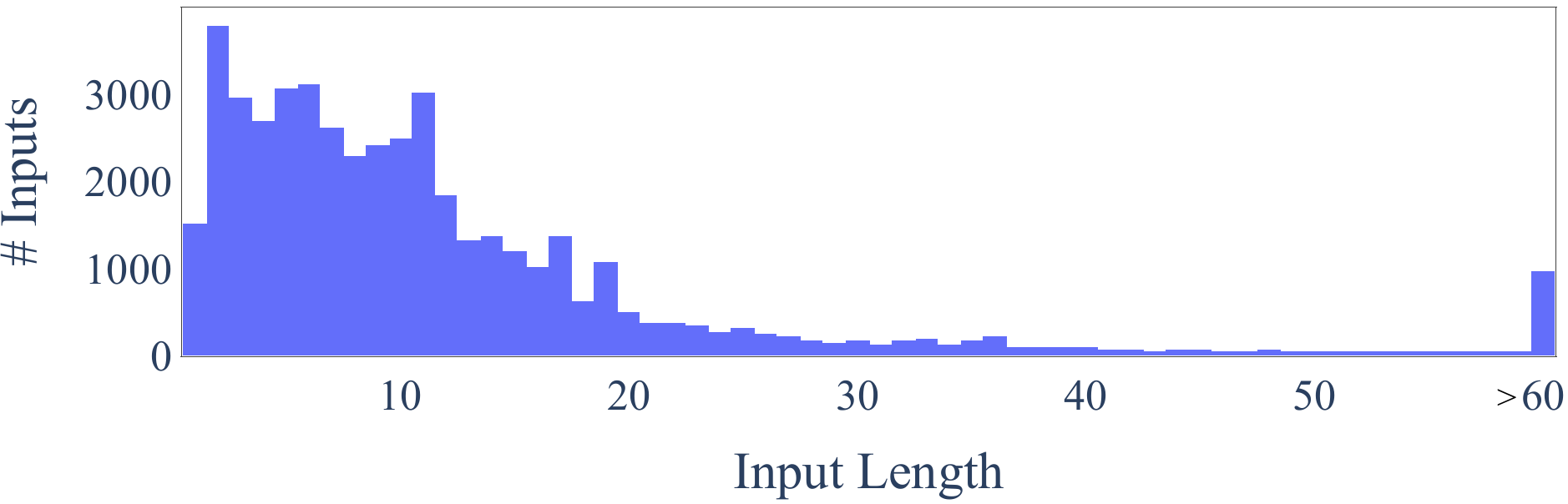}
    \end{subfigure}
    \par\medskip
    \begin{subfigure}{0.99\linewidth}
      \includegraphics[width=\linewidth, trim=0cm 0.45cm 0cm 0cm]{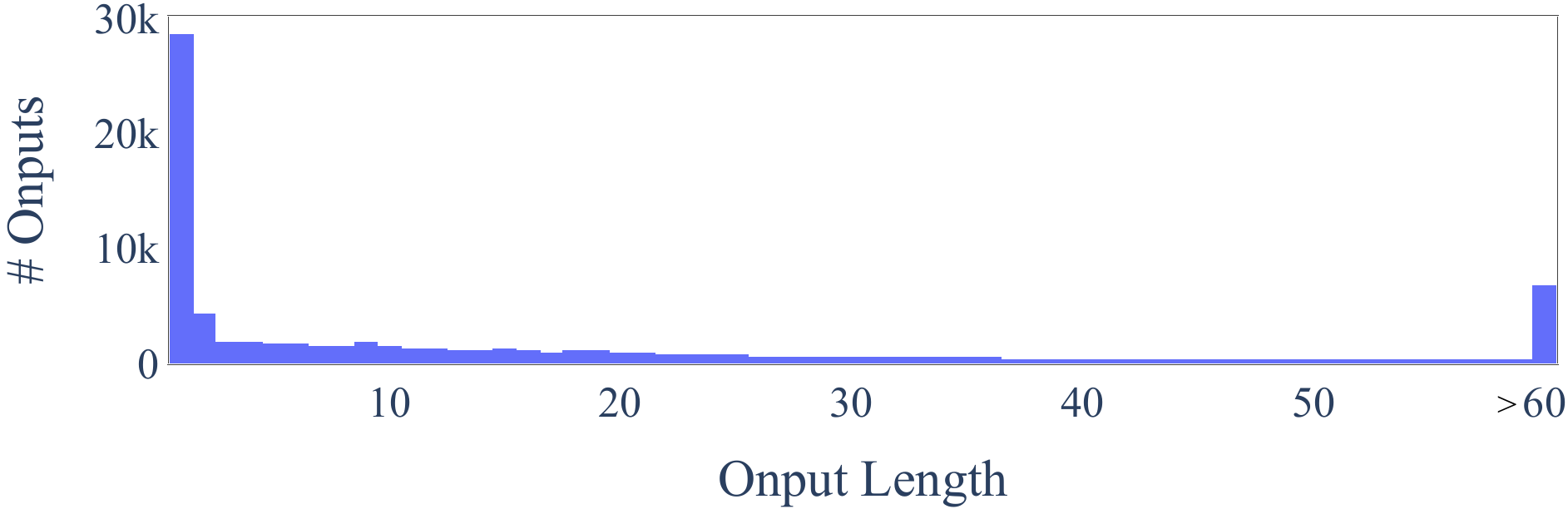}
    \end{subfigure}
    \caption{Length distribution of the generated instructions, non-empty inputs, and outputs.  \label{fig:length_distribution}
  }
  \end{minipage}
  }

\end{figure*}

%% file: tables/data_statistics.tex
\begin{table}[ht]
    \centering
    \small
    \begin{tabular}{lc}
    \hline
    \toprule
    statistic &  \\ 
    \midrule
    \# of instructions & 52,445 \\
    ~~- \# of classification instructions & 11,584 \\
    ~~- \# of non-classification instructions & 40,861 \\
    \# of instances & 82,439 \\
    ~~- \# of instances with empty input & 35,878 \\
    ave.~instruction length (in words) & 15.9 \\
    ave.~non-empty input length (in words) & 12.7 \\
    ave.~output length (in words) &  18.9 \\
    \bottomrule
    \end{tabular}
    \caption{
        Statistics of the generated data by applying \name{} to GPT3.
    }
    \label{tab:data_statistics}
\end{table}

%% file: tables/synthetic_data_quality.tex
\begin{table}[h]
\small
  \setlength\doublerulesep{\arrayrulewidth} 
 \renewcommand\cellset{\renewcommand\arraystretch{0.5}}
\centering
\begin{tabular}{|c|c|}
\toprule
Quality Review Question                                   & Yes \% \\ 
\midrule \midrule
\makecell[c]{Does the instruction \\ describe a valid task? }               &   92\%  \\ \midrule
\makecell[c]{Is the input appropriate \\ for the instruction?}               &   79\% \\ \midrule
\makecell[c]{Is the output a correct and acceptable \\ response to the instruction and input?} &   58\%  \\ \midrule
\midrule
All fields are valid                                        &  54\%   \\ \bottomrule
\end{tabular}
\caption{Data quality review for the instruction, input, and output of the generated data. See \autoref{tab:generated_tasks} and \autoref{tab:bad-generated-tasks} for representative valid and invalid examples.
}
\label{tab:data_quality_eval}
\end{table}

%% file: tables/superni_results.tex
\newcommand{\xbiasAgain}{-54.5ex}
\newcommand{\ybiasAgain}{2ex}

\begin{table}[t]
\small
\hfill
{
\renewcommand{\arraystretch}{1.1}%
\setlength\tabcolsep{1.5pt} 
\resizebox{0.95\linewidth}{!}{
\begin{tabular}{lcc}
\toprule
\textbf{Model}                                                   & \textbf{\# Params}            & \textbf{ROUGE-L} \\
\midrule
\textbf{Vanilla LMs}                                                    \\
T5-LM                                                            & 11B                           & 25.7                                            \\
\tikzmark{a} {\color{black} \gptthree{}}                                                            & 175B                          & 6.8                                                 \\
\midrule

\textbf{Instruction-tuned w/o {\color{black}\supernatShort}}                    \\
\tzero                                                               & 11B                           & 33.1                                                \\
{\color{black} \gptthree{}} + \tzero{} Training                                               & 175B                          & 37.9                                                \\
\tikzmark{b}\tikzmark{c}{\color{black} \gptself{} }
(Ours)                            & 175B                          & 39.9                                                \\
\tikzmark{d}{\color{black}\gptinstruct{001}}                                   & 175B                          & \textbf{40.8}                                                \\
\midrule

\textbf{Instruction-tuned w/ {\color{black}\supernatShort}}             \\
\tkinstruct{}                                                      & 11B                           & 46.0                                                  \\
\tikzmark{f}{\color{black} \gptthree{}} + {\color{black}\supernatShort{}} Training                                          & 175B                          & 49.5                                                \\
\tikzmark{e}{\color{black} \gptself{}}
+ {\color{black}\supernatShort{}} Training   (Ours)       & 175B                          & \textbf{51.6} \\
\bottomrule
\end{tabular}
}
\begin{tikzpicture}[ remember picture, overlay]
    \draw [<->]  ([yshift=1.2ex,xshift=-0.5ex]pic cs:b) [bend left=30] to ([yshift=-0.6ex,xshift=-0.5ex, looseness=2]pic cs:a);
    \node at (\xbiasAgain,0.8ex + \ybiasAgain) {\textcircled{\tiny 1}};
    \draw [<->]  ([yshift=0.95ex,xshift=-0.5ex]pic cs:d) [bend left=60] to ([yshift=0.95ex,xshift=-0.5ex, looseness=2]pic cs:c);
    \node at (\xbiasAgain,-5.9ex +\ybiasAgain) {\textcircled{\tiny 2}};
    \draw [<->]  ([yshift=3ex,xshift=-0.5ex]pic cs:e) [bend left=90] to ([yshift=3.2ex,xshift=-0.5ex, looseness=2]pic cs:f);
    \node at (\xbiasAgain,-16ex+\ybiasAgain) {\textcircled{\tiny 3}};
\end{tikzpicture}
\caption{Evaluation results on \emph{unseen} tasks from {\color{black}\supernatShort{}} (\S\ref{subsec:superni-experiments}). 
From the results, we see that \textcircled{\tiny 1}  \name{} can boost \gptthree{} performance by a large margin (+33.1\%) and \textcircled{\tiny 2} nearly matches the performance of \gptinstruct{001}. 
Additionally, \textcircled{\tiny 3} it can further improve the performance even when a large amount of labeled instruction data is present. 
}
\label{tab:superni_results}
}
\end{table}

%% file: tables/hyper_parameters.tex
\setlength\tabcolsep{2 pt} 
\begin{table*}[h]
\centering
\small
\begin{tabular}{|l|c|c|c|c|c|c|c|}
\toprule
  Experiments $\downarrow$                               & Temp. & Top\_P & Freq. Penalty & Presence Penalty & Beam Size & Max Length & Stop Sequences                                                             \\ \midrule
Generating instructions          & 0.7        & 0.5    & 0                 & 2                & 1         & 1024                  & "\textbackslash{}n\textbackslash{}n", "\textbackslash{}n16", "16.", "16 ." \\
Identifying clf. tasks & 0           & 0      & 0                 & 0                & 1         & 3                     & "\textbackslash{}n", "Task:"                                               \\
Generating instances             & 0           & 0      & 0                 & 1.5              & 1         & 300                   & "Task:"                                                                    \\
Evaluating models    & 0           & 0      & 0                 & 0                & 0         & 1024                  & None (default) \\
\bottomrule
\end{tabular}
    \caption{Hyper-parameters for querying OpenAI API in different experiments.}
    \label{tab:query-gpt3-parameters}
\end{table*}

%% file: tables/instruction_generation_template.tex



\newcommand{\pa}[1]{ {\color{blue}\{} #1 {\color{blue}\}} }

\begin{table*}[ht]
    \centering
    \small
    \noindent\fbox{%
    \begin{minipage}{\dimexpr\linewidth-200\fboxsep-2\fboxrule} 
\tt 
Come up with a series of tasks:\\
\\
Task 1: \{instruction for existing task 1\} \\
Task 2: \{instruction for existing task 2\} \\
Task 3: \{instruction for existing task 3\} \\
Task 4: \{instruction for existing task 4\} \\
Task 5: \{instruction for existing task 5\} \\
Task 6: \{instruction for existing task 6\} \\
Task 7: \{instruction for existing task 7\} \\
Task 8: \{instruction for existing task 8\} \\
Task 9:
    \end{minipage}
}

    \caption{Prompt used for generating new instructions. 8 existing instructions are randomly sampled from the task pool for in-context demonstration. The model is allowed to generate instructions for new tasks, until it stops its generation, reaches its length limit or generates ``Task 16'' tokens.}
    \label{tab:instruction_generation_template}
\end{table*}

%% file: tables/classification_task_or_not_template.tex
\begin{table*}[]
    \centering
    \small
    \noindent\fbox{%
    \begin{minipage}{\dimexpr\linewidth-2\fboxsep-2\fboxrule} 
\tt
Can the following task be regarded as a classification task with finite output labels?\\
\\
Task: Given my personality and the job, tell me if I would be suitable.\\
Is it classification? Yes\\
\\
Task: Give me an example of a time when you had to use your sense of humor.\\
Is it classification? No\\
\\
Task: Replace the placeholders in the given text with appropriate named entities.\\
Is it classification? No\\
\\
Task: Fact checking - tell me if the statement is true, false, or unknown, based on your knowledge and common sense.\\
Is it classification? Yes\\
\\
Task: Return the SSN number for the person.\\
Is it classification? No\\
\\
Task: Detect if the Reddit thread contains hate speech.\\
Is it classification? Yes\\
\\
Task: Analyze the sentences below to identify biases.\\
Is it classification? No\\
\\
Task: Select the longest sentence in terms of the number of words in the paragraph, output the sentence index.\\
Is it classification? Yes\\
\\
Task: Find out the toxic word or phrase in the sentence.\\
Is it classification? No\\
\\
Task: Rank these countries by their population.\\
Is it classification? No\\
\\
Task: You are provided with a news article, and you need to identify all the categories that this article belongs to. Possible categories include: Music, Sports, Politics, Tech, Finance, Basketball, Soccer, Tennis, Entertainment, Digital Game, World News. Output its categories one by one, seperated by comma.\\
Is it classification? Yes\\
\\
Task: Given the name of an exercise, explain how to do it.\\
Is it classification? No\\
\\
Task: Select the oldest person from the list.\\
Is it classification? Yes\\
\\
Task: Find the four smallest perfect numbers.\\
Is it classification? No\\
\\
Task: Does the information in the document supports the claim? You can answer "Support" or "Unsupport".\\
Is it classification? Yes\\
\\
Task: Create a detailed budget for the given hypothetical trip.\\
Is it classification? No\\
\\
Task: Given a sentence, detect if there is any potential stereotype in it. If so, you should explain the stereotype. Else, output no.\\
Is it classification? No\\
\\
$\cdots$ \\
\\
Task: To make the pairs have the same analogy, write the fourth word.\\
Is it classification? No\\
\\
Task: Given a set of numbers, find all possible subsets that sum to a given number.\\
Is it classification? No\\
\\
Task: \{instruction for the target task\}
    \end{minipage}
}

    \caption{Prompt used for classifying whether a task instruction is a classification task or not.}
    \label{tab:classification_task_identification_template}
\end{table*}

%% file: tables/input_first_template.tex
\begin{table*}[]
    \centering
    \small
    \noindent\fbox{%
    \begin{minipage}{\dimexpr\linewidth-2\fboxsep-2\fboxrule}
\tt
Come up with examples for the following tasks. Try to generate multiple examples when possible. If the task doesn't require additional input, you can generate the output directly.\\
\\
Task: Which exercises are best for reducing belly fat at home?\\
Output:\\
- Lying Leg Raises\\
- Leg In And Out\\
- Plank\\
- Side Plank\\
- Sit-ups\\
\\
Task: Extract all the country names in the paragraph, list them separated by commas.\\
Example 1\\
Paragraph: Dr. No is the sixth novel by the English author Ian Fleming to feature his British Secret Service agent James Bond. Written at Fleming's Goldeneye estate in Jamaica, it was first published in the United Kingdom by Jonathan Cape in 1958. In the novel Bond looks into the disappearance in Jamaica of two fellow MI6 operatives who had been investigating Doctor No. Bond travels to No's Caribbean island and meets Honeychile Rider, who is there to collect shells. They are captured and taken to a luxurious facility carved into a mountain. The character of Doctor No, the son of a German missionary and a Chinese woman, was influenced by Sax Rohmer's Fu Manchu stories. Dr. No was the first of Fleming's novels to face widespread negative reviews in Britain, but it was received more favourably in the United States.\\
Output: English, British, Jamaica, the United Kingdom, German, Chinese, Britain, the United States.\\
\\
Task: Converting 85 F to Celsius.\\
Output: 85°F = 29.44°C\\
\\
Task: Sort the given list ascendingly. \\
Example 1\\
List: [10, 92, 2, 5, -4, 92, 5, 101]\\
Output: [-4, 2, 5, 5, 10, 92, 92, 101]\\
Example 2\\
Input 2 - List: [9.99, 10, -5, -1000, 5e6, 999]\\
Output: [-1000, -5, 9.99, 10, 999, 5e6]\\
\\
Task: Suggest a better and more professional rephrasing of the following sentence.\\
Example 1\\
Sentence: This house is surprisingly not constructed very well, and you probably need more money to fix it after you buy it. If you ask me, I would suggest you to consider other candidates.\\
Output: This house does not seem to be constructed well, so you may need to spend more money to fix it after you purchase it. I would suggest that you look at other properties.\\
Example 2\\
Sentence: Just so you know, we did an experiment last week and found really surprising results - language model can improve itself!\\
Output: Our experiments last week demonstrated surprising results, proving that the language model can improve itself.\\
\\
$\cdots$ \\
\\
Task: Turn down a job offer by sending an email to a recruiter explaining the reason.\\
Output: Hi  [Recruiter],\\
Thank you so much for the generous offer to join your team. As we discussed, I’ve admired the company for a number of years, and am a proud endorser of its products. However, after further consideration of where I currently am in my career, I’ve decided to accept an offer at another company.\\
I would love to stay in touch with you and have already started following you on [Social Media Platform]. Again, thank you so much for your time and consideration.\\
Thanks again,\\
\lbrack Your Name\rbrack \\\
\\
Task: \{Instruction for the target task\}
    \end{minipage}
}
    \caption{Prompt used for the input-first approach of instance generation. The model is prompted to generate the instance first, and then generate the corresponding output. For instructions that don't require additional input, the output is allowed to be generated directly.}
    \label{tab:input-first-generation-template}
\end{table*}

%% file: tables/output_first_template.tex
\begin{table*}[]
    \centering
    \small
    \noindent\fbox{%
    \begin{minipage}{\dimexpr\linewidth-2\fboxsep-2\fboxrule}
\tt
Given the classification task definition and the class labels, generate an input that corresponds to each of the class labels. If the task doesn't require input, just generate the correct class label.\\
\\
Task: Classify the sentiment of the sentence into positive, negative, or mixed.\\
Class label: mixed\\
Sentence: I enjoy the flavor of the restaurant but their service is too slow.\\
Class label: Positive\\
Sentence: I had a great day today. The weather was beautiful and I spent time with friends.\\
Class label: Negative\\
Sentence: I was really disappointed by the latest superhero movie. I would not recommend it.\\
\\
Task: Given a dialogue, classify whether the user is satisfied with the service. You should respond with "Satisfied" or "Unsatisfied".\\
Class label: Satisfied\\
Dialogue:\\
- Agent: Thank you for your feedback. We will work to improve our service in the future.\\
- Customer: I am happy with the service you provided. Thank you for your help.\\
Class label: Unsatisfied\\
Dialogue:\\
- Agent: Sorry that we will cancel your order. You will get a refund within 7 business days.\\
- Customer: oh that takes too long. I want you to take quicker action on this.\\
\\
Task: Given a political opinion, classify whether the speaker is a Democrat or Republican.\\
Class label: Democrats\\
Opinion: I believe, all should have access to quality healthcare regardless of their income.\\
Class label: Republicans\\
Opinion: I believe that people should be able to keep more of their hard-earned money and should not be taxed at high rates.\\
\\
Task: Tell me if the following email is a promotion email or not.\\
Class label: Promotion\\
Email: Check out our amazing new sale! We've got discounts on all of your favorite products.\\
Class label: Not Promotion\\
Email: We hope you are doing well. Let us know if you need any help.\\
\\
Task: Detect if the Reddit thread contains hate speech.\\
Class label: Hate Speech\\
Thread: All people of color are stupid and should not be allowed to vote.\\
Class label: Not Hate Speech\\
Thread: The best way to cook a steak on the grill.\\
\\
Task:  Does the document supports the claim? Answer with "Support" or "Unsupport".\\
Class label: Unsupport\\
Document: After a record-breaking run that saw mortgage rates plunge to all-time lows and home prices soar to new highs, the U.S. housing market finally is slowing. While demand and price gains are cooling, any correction is likely to be a modest one, housing economists and analysts say. No one expects price drops on the scale of the declines experienced during the Great Recession.\\
Claim: The US housing market is going to crash soon.\\
Class label: Support\\
Document: The U.S. housing market is showing signs of strain, with home sales and prices slowing in many areas. Mortgage rates have risen sharply in recent months, and the number of homes for sale is increasing. This could be the beginning of a larger downturn, with some economists predicting a potential housing crash in the near future.\\
Claim: The US housing market is going to crash soon.\\
\\
$\cdots$ \\
\\
Task: Which of the following is not an input type? (a) number (b) date (c) phone number (d) email address (e) all of these are valid inputs.\\
Class label: (e)\\
\\
Task: \{instruction for the target task\}
\end{minipage}
}

    \caption{Prompt used for the output-first approach of instance generation. The model is prompted to generate the class label first, and then generate the corresponding input. This prompt is used for generating the instances for classification tasks.}
    \label{tab:output-first-generation-template}
\end{table*}

%% file: tables/case_study.tex
\setlength\tabcolsep{1 pt} 
\newcommand{\sep}{-0.1cm} 
\newcommand{\sty}{\tt \scriptsize}
\renewcommand\labelitemi{--} 

\newcommand{\ins}{{\greentext{Instruction:}}}
\newcommand{\inp}{\newline{\greentext{Input:}}}

\begin{table*}[th!]
\centering
\small
\resizebox{0.99\textwidth}{!}{
\begin{tabular}{p{0.31\textwidth}|p{0.40\textwidth}|p{0.32\textwidth}|c}
\toprule
\multicolumn{1}{c}{\textbf{Instruction + Instance Input}}   & \multicolumn{1}{c}{\textbf{Target Response}} & \multicolumn{1}{c}{\textbf{\gptself{} Response}}   & \multicolumn{1}{c}{\textbf{Rating}}\\
\midrule
\sty
\ins{} Give students tips on how to keep their nerves under control during class presentations. &  
\sty
Everyone gets a little nervous when speaking in front of a group. That’s normal. But the more preparation you’ve done—meaning plenty of researching, organizing, and rehearsing—the more confident you’ll be. Preparation is the key. And if you make a mistake or stumble over your words, just regroup and keep going. Nobody’s perfect, and nobody expects you to be. & 
\vspace{-2mm}
\sty
\begin{itemize}[leftmargin=0.3cm, itemsep=\sep, topsep=0.1pt]
    \item Practice your presentation in front of friends and family. 
    \item Don't drink too much caffeine before the presentation. 
    \item Take deep breaths before you start speaking. 
    \item If you feel nervous, smile. 
    \item Remember that everyone in the audience is rooting for you. 
\end{itemize} & \sty A
\\
\midrule
\sty
\ins{} Make a list of the pros and cons of the given decision. 
\inp{} Implementing a Remote Working Policy 
& 
\sty
Pros: \begin{itemize}[leftmargin=0.3cm, itemsep=\sep, topsep=0.1pt]
    \item Allowing employees to work from home can reduce their stressful commutes and increase their productivity.
    \item Remote working provides greater flexibility, particularly for working parents and caregivers who have other responsibilities. 
    \item \dots (more points)
\end{itemize}

Cons:
\begin{itemize}[leftmargin=0.3cm, itemsep=\sep, topsep=0.1pt]
    \item It may be more difficult for team members to build meaningful, productive, creative relationships with one another.    
    \item \dots (more points)
    \end{itemize}
    &  
    \sty
    Pros:
\begin{itemize}[leftmargin=0.3cm, itemsep=\sep, topsep=0.1pt]
    \item Employees can work from home and save on commuting costs.     
    \item Employees can spend more time with their families.    
\end{itemize}
Cons:
\begin{itemize}[leftmargin=0.3cm, itemsep=\sep, topsep=0.1pt]   
    \item Employees may not be as productive as they would be in the office.    
    \item Employees may not be as motivated to work. 
\end{itemize} & \sty A
\\
\midrule
\sty
\ins{} A user story is an informal, general explanation of a software feature written from the perspective of the end-user or customer. Write a user story for a given software. \inp{} Gmail &  
\sty As a user, I want to be able to recover the password to my account, so that I will be able to access my account in case I forgot the password.
 & 
\sty As a user, I want to be able to send emails to my friends and family. I want to be able to attach files to my emails. I want to be able to search for emails in my inbox. & \sty A
\\
\midrule
\sty
\ins{} A color description has been provided. Find the CSS code associated with that color. 
\inp{} A light red color with a medium light shade of pink & 
\sty
\#DB7093 ($\textcolor[HTML]{DB7093}\lgblksquare$)& 
\sty
~\#ff66cc ($\textcolor[HTML]{ff66cc}\lgblksquare$) & \sty B \\
\midrule

\sty
\ins{} Someone gives you some information about where they want to go and what they enjoy seeing and doing. Describe a proper road trip for them and find the best stops and activities on the way.  
\inp{} I'm looking for a west coast road trip. My goal is to start near San Diego and travel to Los Angeles. &
\sty
Begin near the city of San Diego and make the first stop in coastal La Jolla. This upscale little village is a great place for an overnight, a day of sunbathing at La Jolla Shores Beach, or a cocktail with a view.
Just 20 minutes north, stop into beachy Encinitas to window shop the boutiques or sample fish tacos. Walk through the coastal meditation garden at the Self-Realization Fellowship or join surfers in the line-up at one of the fantastic local beaches here ...(more ideas)
&  
\sty  You should start your trip in San Diego, California. You can visit the San Diego Zoo, which is one of the best zoos in the world. Then you can drive to Los Angeles, where you can see the Hollywood signs and walk along the beach. & \sty B\\
\midrule
\sty
\ins{} Give some examples of what people usually say in the given social situation.  
\inp{} when someone arrives safely &
\sty
\vspace{-2mm}
\begin{itemize}[leftmargin=0.3cm, itemsep=\sep, topsep=0.1pt]
    \item ``Glad you made it safe and sound.''
    \item ``Good to see you got here unscathed.''
    \item ``So, bulletproof vests actually work!''
\end{itemize} 
&  
\sty
Thank God, I made it. & \sty B\\
\midrule
\sty
\ins{} You are given a sentence in first-order logic. Translate it into natural language.  
\inp{}
$\exists x \forall	y (soldier(x) $ 
$\to general(x,y))$ 
& 
\sty
There is someone (x) such that if he is a soldier, then he is general of everyone. &  
\sty
There exists a soldier such that for every general, he is a general. & \sty C\\
\midrule 
\sty 
\ins{} Make a questionnaire to help hotel guests write hotel reviews. & 
\sty \begin{itemize}[leftmargin=0.3cm, itemsep=\sep, topsep=0 pt]
\item Your overall rating of this hotel:
\item Summarize your visit or highlight an interesting detail:
\item ...(more questions)
 \end{itemize}&
\sty  Question: How would you rate the service at the front desk? \newline
Answer: Excellent, Good, Average, Poor
& \sty C
 \\
\midrule
\sty 
\ins{}
Design an outline for a blog post based on the given information and list the sections accordingly. 
\inp{}
 This blog post is going to about making a list of last-minute gift ideas. Also, the tone of the post is going to be relaxed and casual. & 
\sty \begin{itemize}[leftmargin=0.3cm, itemsep=\sep, topsep=0 pt]
\item Section: Movie tickets
\item Section: A fun game console
\item Section: Personalized items with photos/artwork
\item ...(more sections)
\item Takeaway: Don't stress about out running out of time to buy, make a gift. \end{itemize}&
\sty \begin{itemize}[leftmargin=0.3cm, itemsep=\sep, topsep=0 pt]
\item Introduction 
\item List of Gift Ideas
\item Conclusion \end{itemize}
& \sty C
 \\
\midrule

\sty
\ins{}
Write a Jira ticket for the given task. 
\inp{}
New Employee onboarding & 
\sty
Title: Please help our new employee get settled.
Prepare a desk, chair, and computer for them to use.
Create their company account, email, and temporary password.
Grant access to company documents. & 
\sty
Hi [Manager],
I’m writing to you because I have a question about the new employee onboarding process. I’ve been working here for a few months now, and I’m still not sure how it works. Could you please explain it to me? & \sty D\\
\midrule
\sty 
\ins{}
Create alliterations by finding synonyms for words in the given sentence. 
\inp{}
David wears a hat everyday. & \sty David dons a derby daily. & \sty  David wears a hat every day. & \sty D \\
\bottomrule
\end{tabular}
}
\caption{Examples in the user-oriented instructions dataset (\S\ref{sec:user_instructions}) and predictions from \gptself{}. 
The right column indicates one of the four quality ratings assigned to the model's response, with ``A'' indicating  ``valid and satisfying'' responses (highest)
and  ``D'' indicating ``irrelevant or invalid response'' (lowest).
}
\label{tab:case_study}
\end{table*}

%% file: tables/examples.tex
\setlength\tabcolsep{1 pt} 
\renewcommand{\sep}{-0.1cm} 
\renewcommand{\sty}{\tt \scriptsize}
\renewcommand\labelitemi{--} 

\renewcommand{\ins}{{\greentext{Instruction:}}}
\renewcommand{\inp}{\newline{\greentext{Input:}}}

\newcommand{\labelssize}{\scriptsize}
\setlength\tabcolsep{4 pt} 
\renewcommand\labelitemi{--} 
\newcommand\tab[1][0.5cm]{\hspace*{#1}}

\newcommand{\allvalid}{\labelssize
Instruction: \cmark \newline Input: \cmark \newline Output: \cmark}
\newcommand{\out}{\newline{\greentext{Target Output:}}}

\begin{table*}[th!]
\centering
\small
\resizebox{1\textwidth}{!}{
\begin{tabular}{p{0.35\textwidth}|p{0.56\textwidth} |p{0.09\textwidth}}
\toprule
\multicolumn{1}{c}{\textbf{Instruction + Instance Input}}  & \multicolumn{1}{c}{\textbf{Target Output}} & \textbf{Validity}\\
\midrule
\sty 
\ins{}
Generate a random password with at least 6 characters.
   & 
\sty def generateRandomPassword():
\newline \tab   password = ""
\newline \tab    while len(password) < 6:
\newline \tab \tab password += chr(random.randint(65, 90))
\newline \tab    return password 
  & \allvalid
\\
\midrule
\sty 
\ins{} Write a paragraph describing how to tie shoelaces for people who have never tied their shoes before. explain each step briefly.
& 
\sty Paragraph: To tie shoelaces, first you need to make a loop with one end of the lace. Then pass the other end through the loop and pull it tight. Next, you need to wrap the lace around the shoe several times. Finally, tuck in the loose end into the wrapped part of the lace.&
\allvalid \\ 
\midrule
\sty 
\ins{}
Given a word, find out its length and its number of vowels. 
\inp{}
 Word = "hello" & \sty Length = 5, Number of vowels = 2 
 & \allvalid
 \\
\midrule
\sty
\ins{}
Write a story with three characters: a person, an animal and an object. 
\inp{}
Person: John, a programmer. \newline
Animal: A dog. \newline
Object: A laptop. \newline &
\sty John was sitting on the floor in front of his laptop when he heard a knock at the door. He opened it to find a dog standing there. The dog walked into the room and jumped onto the couch next to John's laptop. & \allvalid\\
\midrule
\sty 
\ins{}
Compose an email and send it to your friend, asking for advice on what to do in this situation. 
\inp{}
Situation: You are a programmer and you have been working on a project for the past few months. The deadline is approaching, but you haven't finished yet. Your manager has asked you to stay late every day this week to finish it. & \sty
Hi [Friend], \newline
I’m having a bit of trouble with my current job. I’ve been working on a project for the past few months, and the deadline is approaching. My manager has asked me to stay late every day this week to finish it. I don’t think I can do that because I have other commitments. What should I do?
&
\allvalid \\ 
\midrule
\sty 
\ins{}
What are some ways we can make our school more eco-friendly?  & \sty
\begin{itemize}[leftmargin=0.3cm, itemsep=\sep, topsep=0.1pt]
\item Use reusable water bottles instead of disposable plastic ones.
\item Turn off lights when you leave a room.
\item Recycle paper, cans and bottles.
\item Plant trees to absorb carbon dioxide.
\end{itemize} & \allvalid \\
\midrule
\sty \ins{} I am looking for a new apartment. there are many options available but they are all quite expensive. i want to move as soon as possible. give me advice on what to do.  & \sty You should look for a cheaper apartment. You can try to negotiate with the landlord and offer him/her a lower price. If you are lucky, they will accept your offer. Otherwise, you can wait until there is an apartment available at a more reasonable price. 
& 
\allvalid \\ 
\midrule
\sty 
\ins{}
Write a dialogue between two people who have been friends for a long time, and one of them is trying to convince the other to do something that he or she doesn't want to do. & \sty
A: Hey, I have a great idea! Let's go to the beach today.\newline
B: No thanks, I'm not really in the mood for that.\newline
A: Come on, it'll be fun! We haven't been there together in ages.
B: I know, but I just don't feel like going.\newline
A: But you love the beach! You always say how much you enjoy swimming and lying in the sun.\newline
B: Yeah, I do. It's just that I've been feeling kind of tired lately.\newline
A: Oh, come on. You're never tired. You're always so energetic.\newline
B: Well, I am now. I think I need some rest.\newline
A: Okay, fine. If you change your mind, let me know.
& 
\allvalid
\\
\midrule
\sty 
\ins{} How can i create an app that will be able to identify different kinds of plants by taking photos of them?
& \sty You can use the following steps to create an app that will be able to identify different kinds of plants by taking photos of them. \newline
Step 1: Create a database with all the information about the plants you want to identify. \newline
Step 2: Create a user interface for your application. \newline
Step 3: Use image recognition techniques to identify the plant in the photo. \newline
Step 4: Display the name of the plant on the screen.
& \allvalid \\
\midrule
\sty 
\ins{} Find the closest matching emoji to a given one. \inp{} Emoji: \includegraphics[scale=0.004]{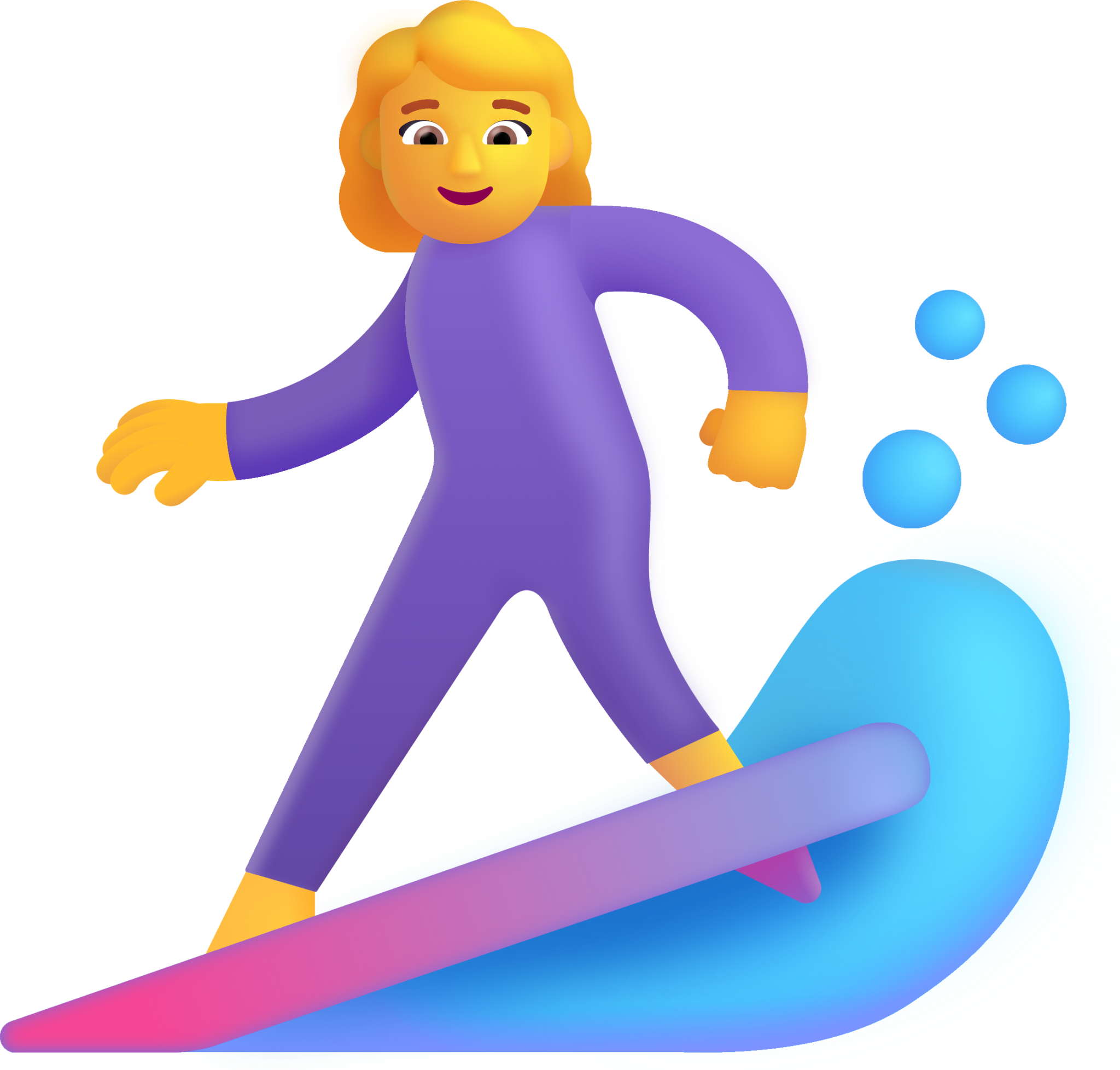}
& \sty \includegraphics[scale=0.004]{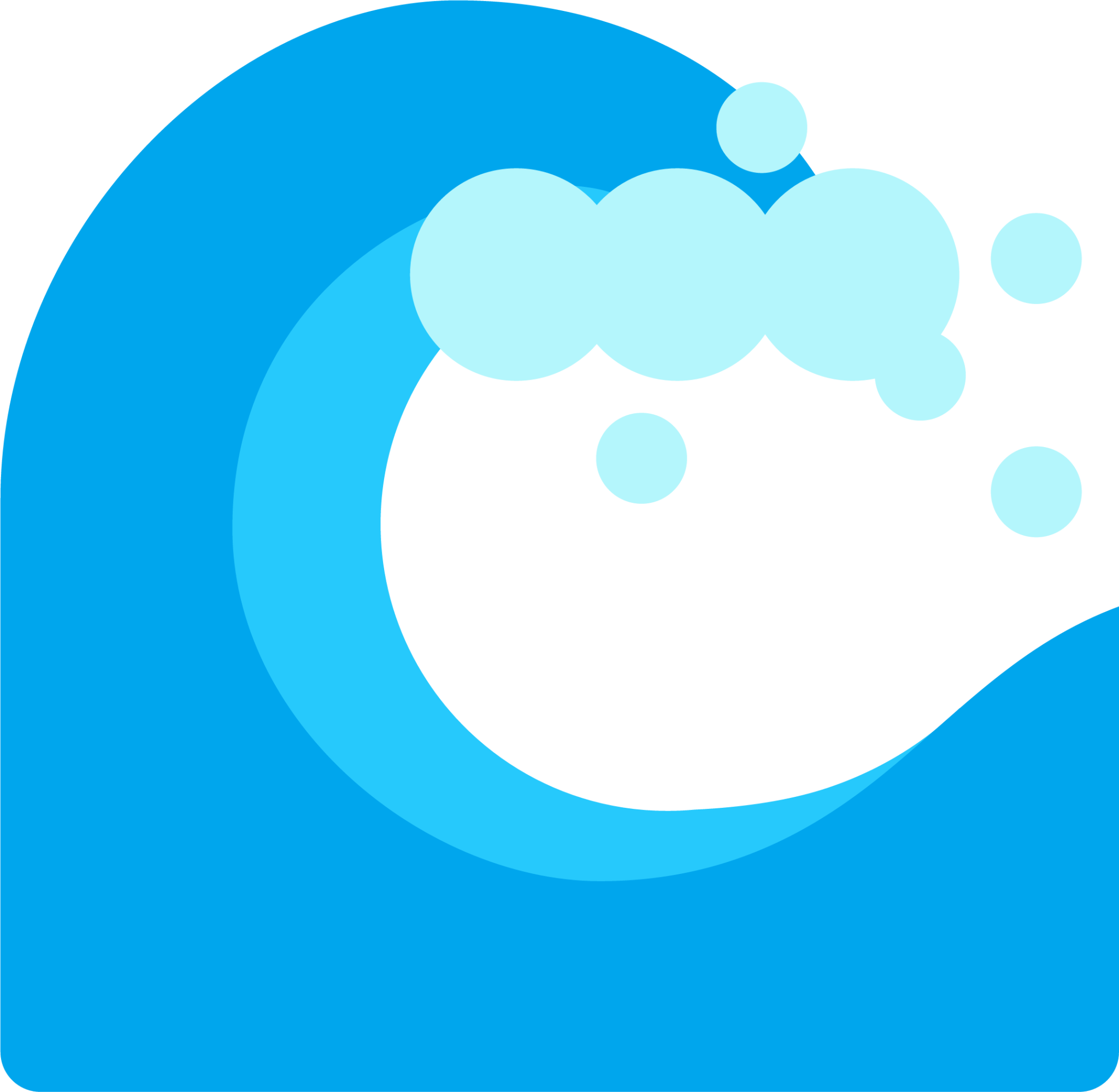}
& \allvalid \\
\midrule
\sty 
\ins{} Describe your favorite video game using only verbs. \inp{} Game: The Legend of Zelda
& \sty Explore, fight, solve, upgrade, repeat
& \allvalid \\
\bottomrule
\end{tabular}
}
\caption{Representative \textbf{valid} tasks generated by GPT3. As is discussed in \S\ref{sec:generatd-data}, these generated tasks cover a broad range of formats, text types, and underlying expertise, while being correct on more than half of all the generated tasks. 
}
\label{tab:generated_tasks}
\end{table*}

%% file: tables/error_examples.tex
\setlength\tabcolsep{3 pt} 
\renewcommand\labelitemi{--} 

\begin{table*}[th!]
\centering
\small
\resizebox{1\textwidth}{!}{
\begin{tabular}{p{0.35\textwidth}|p{0.56\textwidth} |p{0.09\textwidth}}
\toprule
\multicolumn{1}{c}{\textbf{Instruction + Instance Input}}  & \multicolumn{1}{c}{\textbf{Target Output}} & \textbf{Validity}\\
\midrule
\sty 
\ins{} 
Given a set of words $w=\{	w_1, w_2 ,..., w_n\}$ where each word $w_i$ has $n$ letters, write a function that takes a target length $l$ and returns a new set of words $s$ such that $|s| = l$ and $|\{i: w_i \in s\}| \leq l / |w|$. &  
\sty def wordSubsetSum(w, l):
\newline \tab    s = []
\newline \tab    for i in range(len(w)):
\newline \tab \tab       if len(s) + 1 < l:
\newline \tab \tab \tab           s.append(w[i])
\newline \tab   return s& 
\labelssize Instruction: \xmark \newline Input: \cmark \newline Output: \xmark\\
\midrule
\sty
\ins{} 
The house on fire! go! &  
\sty \begin{itemize}[leftmargin=0.3cm, itemsep=\sep, topsep=0.1pt]
\item Go to the kitchen and take a pan.
\item Take water from the tap.
\item Pour water into the pan.
\item Put out the fire with the pan.
\end{itemize}
& 
\labelssize Instruction: \xmark \newline Input: \cmark \newline Output: \xmark\\
\midrule
\sty
\ins{} Find out if you have any friends who like to play football or chess. &  
\sty def hasFriendsWhoLikeToPlayFootballOrChess(self):
\newline \tab friends = []
\newline \tab for friend in self.friends:
\newline \tab \tab if friend.likes == 'football' or friend.likes == 'chess':
\newline \tab \tab friends.append(friend)
\newline \tab return len(friends) > 0
& 
\labelssize Instruction: \xmark \newline Input: \cmark \newline Output: \cmark\\
\midrule
\sty
\ins{} 
A number of people are lined up for an interview. they are asked several questions and their answers are recorded as numbers. you need to identify which person has the highest salary among all the candidates. 
\inp{}
Person 1: 2, 3, 4, 5, 6 \newline
Person 2: 7, 8, 9, 10, 11 \newline
Person 3: 12, 13, 14, 15, 16 \newline
Person 4: 17, 18, 19, 20, 21 \newline
Person 5: 22, 23, 24, 25, 26 \newline
Person 6: 27, 28, 29, 30, 31 & \sty Highest salary &  
\labelssize
Instruction: \cmark \newline Input: \xmark \newline Output: \xmark \\
\midrule
\sty 
\ins{}
Given a list of words, generate a story about a group of people who live on a planet. the story should contain at least 10 sentences. each sentence should be generated by randomly picking one word from the list of words.
   & 
\sty The story starts with a group of people who live on a planet. They are very intelligent and have created many things that help them in their daily lives. One day, they decided to create a machine that can make food for them. The machine was able to produce food but it also produced something else. It produced a liquid that could be used as fuel for the machines. The people were happy because they didn’t need to go out and look for food or fuel anymore. They had everything they needed right there in their homes. However, one day, the machine stopped working. The people tried to fix it but couldn’t figure out what went wrong. They tried to find someone who could help them but no one knew how to fix the machine. After a few days, the people realized that they would die if they didn’t do anything about it. So, they decided to send some of their best scientists to space to find a new planet where they could live. The scientists traveled for a long time before they found a planet that looked like a good place to live. They sent a message back to the people telling them about the planet. The people were excited and started preparing for the journey. They built spaceships and flew to the new planet. When they arrived, they saw that the planet was beautiful. There were trees and rivers and mountains. The people were so happy that they finally found a new home. They started building houses and planting crops. They lived happily ever after. &
\labelssize
Instruction: \cmark \newline Input: \xmark \newline Output: \cmark
\\
\midrule
\sty \ins{} Find out what the author's attitude towards working part-time is. &
\sty Paragraph: Working part-time is a great way to earn some extra money while still being able to enjoy your free time. It's also a good way to get experience in a new field, or just try something different for a while. However, it can be difficult to find the right balance between work and play. \newline
The author seems to be positive about working part-time. &
\labelssize
Instruction: \cmark \newline Input: \xmark \newline Output: \cmark \\
\midrule
\sty \ins{} Is it ethical to use a new technology that could potentially cause harm to others? &
\sty 
New Technology: Self-driving cars \newline
Potential Harm: Accidents caused by self-driving cars \newline
Ethical? Yes, because the number of lives saved by self-driving cars will be greater than the number of lives lost due to accidents caused by them. &
\labelssize
Instruction: \cmark \newline Input: \xmark \newline Output: \cmark \\
\midrule
\sty 
\ins{}
What is your salary? & 
\sty
class Solution(object):
\newline \tab def getSalary(self):
\newline \tab \tab   return self.salary & 
\labelssize
Instruction: \cmark \newline Input: \cmark \newline Output: \xmark\\
\midrule
\sty 
\ins{}
A person wants to buy a car from another person. he has some bitcoins, but he doesn't want to reveal his identity. how can they make the transaction happen? &  
\sty The buyer and seller should agree on a price for the car. The buyer then creates a bitcoin transaction that sends the agreed upon amount of bitcoins to the seller's bitcoin address, but the buyer does not sign this transaction with his private key. Instead, he gives the unsigned transaction to the seller. The seller can now sign the transaction using his private key, and broadcast it to the network. Once the transaction is confirmed by the network, the buyer has no way to get his money back, so he will be forced to go through with the purchase. & 
\labelssize
Instruction:\cmark \newline Input: \cmark \newline Output: \xmark\\
\bottomrule
\end{tabular}
}
\caption{
Representative \textbf{invalid} tasks generated by \gptthree{}. The problematic fields are indicated in the validity column. As discussed in \S\ref{subsec:quality}, although these tasks contain errors, they still provide many useful signals in supervising models to follow instructions.
}
\label{tab:bad-generated-tasks}
\end{table*}

%% file: main.bbl
\providecommand{\CNFX}[1]{{\em{\textrm{(#1)}}}}
\begin{thebibliography}{47}
\expandafter\ifx\csname natexlab\endcsname\relax\def\natexlab#1{#1}\fi

\bibitem[{Amini et~al.(2022)Amini, Feofanov, Pauletto, Devijver, and
  Maximov}]{amini2022self}
Massih-Reza Amini, Vasilii Feofanov, Loic Pauletto, Emilie Devijver, and Yury
  Maximov. 2022.
\newblock \href {https://arxiv.org/abs/2202.12040} {Self-training: A survey}.
\newblock \emph{arXiv preprint arXiv:2202.12040}.

\bibitem[{Bach et~al.(2022)Bach, Sanh, Yong, Webson, Raffel, Nayak, Sharma,
  Kim, Bari, Fevry et~al.}]{bach2022promptsource}
Stephen~H Bach, Victor Sanh, Zheng-Xin Yong, Albert Webson, Colin Raffel,
  Nihal~V Nayak, Abheesht Sharma, Taewoon Kim, M~Saiful Bari, Thibault Fevry,
  et~al. 2022.
\newblock \href {https://arxiv.org/abs/2202.01279} {{PromptSource: An
  Integrated Development Environment and Repository for Natural Language
  Prompts}}.
\newblock In \emph{Annual Meeting of the Association for Computational
  Linguistics \CNFX{ACL} - System Demonstrations}.

\bibitem[{Brown et~al.(2020)Brown, Mann, Ryder, Subbiah, Kaplan, Dhariwal,
  Neelakantan, Shyam, Sastry, Askell, Agarwal, and et~al.}]{brown2020gpt3}
Tom~B. Brown, Benjamin Mann, Nick Ryder, Melanie Subbiah, Jared Kaplan,
  Prafulla Dhariwal, Arvind Neelakantan, Pranav Shyam, Girish Sastry, Amanda
  Askell, Sandhini Agarwal, and et~al. 2020.
\newblock \href
  {https://papers.nips.cc/paper/2020/hash/1457c0d6bfcb4967418bfb8ac142f64a-Abstract.html}
  {{Language models are few-shot learners}}.
\newblock In \emph{Advances in Neural Information Processing Systems
  \CNFX{NeurIPS}}.

\bibitem[{Chung et~al.(2022)Chung, Hou, Longpre, Zoph, Tay, Fedus, Li, Wang,
  Dehghani, Brahma et~al.}]{chung2022scaling}
Hyung~Won Chung, Le~Hou, Shayne Longpre, Barret Zoph, Yi~Tay, William Fedus,
  Eric Li, Xuezhi Wang, Mostafa Dehghani, Siddhartha Brahma, et~al. 2022.
\newblock \href {https://arxiv.org/abs/2210.11416} {Scaling
  instruction-finetuned language models}.
\newblock \emph{arXiv preprint arXiv:2210.11416}.

\bibitem[{Du et~al.(2021)Du, Grave, Gunel, Chaudhary, Celebi, Auli, Stoyanov,
  and Conneau}]{du2021self}
Jingfei Du, {\'E}douard Grave, Beliz Gunel, Vishrav Chaudhary, Onur Celebi,
  Michael Auli, Veselin Stoyanov, and Alexis Conneau. 2021.
\newblock \href {https://aclanthology.org/2021.naacl-main.426} {Self-training
  improves pre-training for natural language understanding}.
\newblock In \emph{Conference of the North American Chapter of the Association
  for Computational Linguistics \CNFX{NAACL}: Human Language Technologies},
  pages 5408--5418.

\bibitem[{Feng et~al.(2021)Feng, Gangal, Wei, Chandar, Vosoughi, Mitamura, and
  Hovy}]{feng2021survey}
Steven~Y Feng, Varun Gangal, Jason Wei, Sarath Chandar, Soroush Vosoughi,
  Teruko Mitamura, and Eduard Hovy. 2021.
\newblock \href {https://aclanthology.org/2021.findings-acl.84/} {A survey of
  data augmentation approaches for nlp}.
\newblock In \emph{Annual Meeting of the Association for Computational
  Linguistics \CNFX{ACL}~ACL-IJCNLP - Findings}, pages 968--988.

\bibitem[{Fried et~al.(2018)Fried, Hu, Cirik, Rohrbach, Andreas, Morency,
  Berg-Kirkpatrick, Saenko, Klein, and Darrell}]{fried2018speaker}
Daniel Fried, Ronghang Hu, Volkan Cirik, Anna Rohrbach, Jacob Andreas,
  Louis-Philippe Morency, Taylor Berg-Kirkpatrick, Kate Saenko, Dan Klein, and
  Trevor Darrell. 2018.
\newblock \href {https://arxiv.org/abs/1806.02724} {Speaker-follower models for
  vision-and-language navigation}.
\newblock In \emph{Advances in Neural Information Processing Systems
  \CNFX{NeurIPS}}.

\bibitem[{He et~al.(2019)He, Gu, Shen, and Ranzato}]{he2019revisiting}
Junxian He, Jiatao Gu, Jiajun Shen, and Marc'Aurelio Ranzato. 2019.
\newblock \href {https://arxiv.org/abs/1909.13788} {Revisiting self-training
  for neural sequence generation}.
\newblock In \emph{International Conference on Learning Representations
  \CNFX{ICLR}}.

\bibitem[{Hinton et~al.(2015)Hinton, Vinyals, Dean
  et~al.}]{hinton2015distilling}
Geoffrey Hinton, Oriol Vinyals, Jeff Dean, et~al. 2015.
\newblock \href {https://arxiv.org/abs/1503.02531} {Distilling the knowledge in
  a neural network}.
\newblock In \emph{Advances in Neural Information Processing Systems
  \CNFX{NeurIPS}~Workshop on Deep Learning}.

\bibitem[{Honovich et~al.(2022{\natexlab{a}})Honovich, Scialom, Levy, and
  Schick}]{honovich2022unnatural}
Or~Honovich, Thomas Scialom, Omer Levy, and Timo Schick. 2022{\natexlab{a}}.
\newblock \href {https://arxiv.org/abs/2212.09689} {Unnatural instructions:
  Tuning language models with (almost) no human labor}.
\newblock \emph{arXiv preprint arXiv:2212.09689}.

\bibitem[{Honovich et~al.(2022{\natexlab{b}})Honovich, Shaham, Bowman, and
  Levy}]{honovich2022instruction}
Or~Honovich, Uri Shaham, Samuel~R Bowman, and Omer Levy. 2022{\natexlab{b}}.
\newblock \href {https://arxiv.org/abs/2205.10782} {Instruction induction: From
  few examples to natural language task descriptions}.
\newblock \emph{arXiv preprint arXiv:2205.10782}.

\bibitem[{Huang et~al.(2022)Huang, Gu, Hou, Wu, Wang, Yu, and
  Han}]{huang2022large}
Jiaxin Huang, Shixiang~Shane Gu, Le~Hou, Yuexin Wu, Xuezhi Wang, Hongkun Yu,
  and Jiawei Han. 2022.
\newblock \href {https://arxiv.org/abs/2205.10782} {Large language models can
  self-improve}.
\newblock \emph{arXiv preprint arXiv:2210.11610}.

\bibitem[{Kandpal et~al.(2022)Kandpal, Deng, Roberts, Wallace, and
  Raffel}]{kandpal2022large}
Nikhil Kandpal, Haikang Deng, Adam Roberts, Eric Wallace, and Colin Raffel.
  2022.
\newblock \href {https://arxiv.org/abs/2211.08411} {Large language models
  struggle to learn long-tail knowledge}.
\newblock \emph{arXiv preprint arXiv:2211.08411}.

\bibitem[{Kitaev et~al.(2019)Kitaev, Cao, and
  Klein}]{kitaev-etal-2019-multilingual}
Nikita Kitaev, Steven Cao, and Dan Klein. 2019.
\newblock \href {https://doi.org/10.18653/v1/P19-1340} {Multilingual
  constituency parsing with self-attention and pre-training}.
\newblock In \emph{Annual Meeting of the Association for Computational
  Linguistics \CNFX{ACL}}, pages 3499--3505.

\bibitem[{Kitaev and Klein(2018)}]{kitaev-klein-2018-constituency}
Nikita Kitaev and Dan Klein. 2018.
\newblock \href {https://doi.org/10.18653/v1/P18-1249} {Constituency parsing
  with a self-attentive encoder}.
\newblock In \emph{Annual Meeting of the Association for Computational
  Linguistics \CNFX{ACL}}, pages 2676--2686.

\bibitem[{Lester et~al.(2021)Lester, Al-Rfou, and Constant}]{lester2021power}
Brian Lester, Rami Al-Rfou, and Noah Constant. 2021.
\newblock \href {https://arxiv.org/abs/2104.08691} {The power of scale for
  parameter-efficient prompt tuning}.
\newblock In \emph{Conference on Empirical Methods in Natural Language
  Processing \CNFX{EMNLP}}.

\bibitem[{Liu et~al.(2022)Liu, Swayamdipta, Smith, and Choi}]{liu2022wanli}
Alisa Liu, Swabha Swayamdipta, Noah~A. Smith, and Yejin Choi. 2022.
\newblock \href
  {https://preview.aclanthology.org/emnlp-22-ingestion/2022.findings-emnlp.508/}
  {{WANLI}: Worker and ai collaboration for natural language inference dataset
  creation}.
\newblock In \emph{Conference on Empirical Methods in Natural Language
  Processing \CNFX{EMNLP} - Findings}.

\bibitem[{Magister et~al.(2022)Magister, Mallinson, Adamek, Malmi, and
  Severyn}]{luciecharlotte2022teachingsmallmodels}
Lucie~Charlotte Magister, Jonathan Mallinson, Jakub Adamek, Eric Malmi, and
  Aliaksei Severyn. 2022.
\newblock \href {https://arxiv.org/abs/2212.08410} {Teaching small language
  models to reason}.
\newblock \emph{arXiv preprint arXiv:2212.08410}.

\bibitem[{Mekala et~al.(2022)Mekala, Vu, Schick, and
  Shang}]{mekala2022intermediate}
Dheeraj Mekala, Tu~Vu, Timo Schick, and Jingbo Shang. 2022.
\newblock \href {https://arxiv.org/abs/2205.12604} {Leveraging qa datasets to
  improve generative data augmentation}.
\newblock \emph{arXiv preprint arXiv:2205.12604}.

\bibitem[{Meng et~al.(2023)Meng, Michalski, Huang, Zhang, Abdelzaher, and
  Han}]{meng2022tuning}
Yu~Meng, Martin Michalski, Jiaxin Huang, Yu~Zhang, Tarek Abdelzaher, and Jiawei
  Han. 2023.
\newblock \href {https://arxiv.org/abs/2211.03044} {Tuning language models as
  training data generators for augmentation-enhanced few-shot learning}.
\newblock In \emph{International Conference on Machine Learning \CNFX{ICML}}.

\bibitem[{Min et~al.(2022)Min, Chaplot, Ravikumar, Bisk, and
  Salakhutdinov}]{min2022film}
So~Yeon Min, Devendra~Singh Chaplot, Pradeep Ravikumar, Yonatan Bisk, and
  Ruslan Salakhutdinov. 2022.
\newblock \href {https://arxiv.org/abs/2110.07342} {{FILM: Following
  Instructions in Language with Modular Methods}}.
\newblock In \emph{International Conference on Learning Representations
  \CNFX{ICLR}}.

\bibitem[{Mishra et~al.(2022)Mishra, Khashabi, Baral, and
  Hajishirzi}]{mishra2022cross}
Swaroop Mishra, Daniel Khashabi, Chitta Baral, and Hannaneh Hajishirzi. 2022.
\newblock \href {https://arxiv.org/abs/2104.08773} {{Cross-Task Generalization
  via Natural Language Crowdsourcing Instructions}}.
\newblock In \emph{Annual Meeting of the Association for Computational
  Linguistics \CNFX{ACL}}.

\bibitem[{Ouyang et~al.(2022)Ouyang, Wu, Jiang, Almeida, Wainwright, Mishkin,
  Zhang, Agarwal, Slama, Ray et~al.}]{ouyang2022training}
Long Ouyang, Jeff Wu, Xu~Jiang, Diogo Almeida, Carroll~L Wainwright, Pamela
  Mishkin, Chong Zhang, Sandhini Agarwal, Katarina Slama, Alex Ray, et~al.
  2022.
\newblock \href {https://arxiv.org/abs/2203.02155} {{Training Language Models
  to Follow Instructions with Human Feedback}}.
\newblock In \emph{Advances in Neural Information Processing Systems
  \CNFX{NeurIPS}}.

\bibitem[{Raffel et~al.(2020)Raffel, Shazeer, Roberts, Lee, Narang, Matena,
  Zhou, Li, and Liu}]{raffel2020exploring}
Colin Raffel, Noam Shazeer, Adam Roberts, Katherine Lee, Sharan Narang, Michael
  Matena, Yanqi Zhou, Wei Li, and Peter~J Liu. 2020.
\newblock \href {https://arxiv.org/abs/1910.10683} {Exploring the limits of
  transfer learning with a unified text-to-text transformer}.
\newblock \emph{Journal of Machine Learning Research \CNFX{JMLR}}.

\bibitem[{Razeghi et~al.(2022)Razeghi, Logan~IV, Gardner, and
  Singh}]{razeghi2022impact}
Yasaman Razeghi, Robert~L Logan~IV, Matt Gardner, and Sameer Singh. 2022.
\newblock \href {https://arxiv.org/abs/2202.07206} {Impact of pretraining term
  frequencies on few-shot reasoning}.
\newblock \emph{arXiv preprint arXiv:2202.07206}.

\bibitem[{Sanh et~al.(2019)Sanh, Debut, Chaumond, and
  Wolf}]{Sanh2019DistilBERTAD}
Victor Sanh, Lysandre Debut, Julien Chaumond, and Thomas Wolf. 2019.
\newblock \href {https://arxiv.org/abs/1910.01108} {Distilbert, a distilled
  version of bert: smaller, faster, cheaper and lighter}.
\newblock In \emph{Advances in Neural Information Processing Systems
  \CNFX{NeurIPS} Workshop on Energy Efficient Machine Learning and Cognitive
  Computing}.

\bibitem[{Sanh et~al.(2022)Sanh, Webson, Raffel, Bach, Sutawika, Alyafeai,
  Chaffin, Stiegler, Raja, Dey, Bari, Xu, Thakker, Sharma, Szczechla, Kim,
  Chhablani, Nayak, Datta, Chang, Jiang, Wang, Manica, Shen, Yong, Pandey,
  Bawden, Wang, Neeraj, Rozen, Sharma, Santilli, Fevry, Fries, Teehan, Scao,
  Biderman, Gao, Wolf, and Rush}]{sanh2022multitask}
Victor Sanh, Albert Webson, Colin Raffel, Stephen Bach, Lintang Sutawika, Zaid
  Alyafeai, Antoine Chaffin, Arnaud Stiegler, Arun Raja, Manan Dey, M~Saiful
  Bari, Canwen Xu, Urmish Thakker, Shanya~Sharma Sharma, Eliza Szczechla,
  Taewoon Kim, Gunjan Chhablani, Nihal Nayak, Debajyoti Datta, Jonathan Chang,
  Mike Tian-Jian Jiang, Han Wang, Matteo Manica, Sheng Shen, Zheng~Xin Yong,
  Harshit Pandey, Rachel Bawden, Thomas Wang, Trishala Neeraj, Jos Rozen,
  Abheesht Sharma, Andrea Santilli, Thibault Fevry, Jason~Alan Fries, Ryan
  Teehan, Teven~Le Scao, Stella Biderman, Leo Gao, Thomas Wolf, and Alexander~M
  Rush. 2022.
\newblock \href {https://arxiv.org/abs/2110.08207} {{Multitask Prompted
  Training Enables Zero-Shot Task Generalization}}.
\newblock In \emph{International Conference on Learning Representations
  \CNFX{ICLR}}.

\bibitem[{Schick and Sch{\"u}tze(2021)}]{schick2021generating}
Timo Schick and Hinrich Sch{\"u}tze. 2021.
\newblock \href {https://aclanthology.org/2021.emnlp-main.555/} {Generating
  datasets with pretrained language models}.
\newblock In \emph{Conference on Empirical Methods in Natural Language
  Processing \CNFX{EMNLP}}.

\bibitem[{Shridhar et~al.(2020)Shridhar, Thomason, Gordon, Bisk, Han, Mottaghi,
  Zettlemoyer, and Fox}]{shridhar2020alfred}
Mohit Shridhar, Jesse Thomason, Daniel Gordon, Yonatan Bisk, Winson Han,
  Roozbeh Mottaghi, Luke Zettlemoyer, and Dieter Fox. 2020.
\newblock \href {https://arxiv.org/abs/1912.01734} {{ALFRED: A Benchmark for
  Interpreting Grounded Instructions for Everyday Tasks}}.
\newblock In \emph{IEEE Conference on Computer Vision and Pattern Recognition
  \CNFX{CVPR}}.

\bibitem[{Singh et~al.(2022)Singh, Morris, Aneja, Rush, and
  Gao}]{singh2022explaining}
Chandan Singh, John~X Morris, Jyoti Aneja, Alexander~M Rush, and Jianfeng Gao.
  2022.
\newblock \href {https://arxiv.org/abs/2210.01848} {Explaining patterns in data
  with language models via interpretable autoprompting}.
\newblock \emph{arXiv preprint arXiv:2210.01848}.

\bibitem[{Sun et~al.(2023)Sun, Shen, Zhou, Zhang, Chen, Cox, Yang, and
  Gan}]{sun2023principle}
Zhiqing Sun, Yikang Shen, Qinhong Zhou, Hongxin Zhang, Zhenfang Chen, David
  Cox, Yiming Yang, and Chuang Gan. 2023.
\newblock \href {https://arxiv.org/abs/2305.03047} {Principle-driven
  self-alignment of language models from scratch with minimal human
  supervision}.
\newblock \emph{arXiv preprint arXiv:2305.03047}.

\bibitem[{Taori et~al.(2023)Taori, Gulrajani, Zhang, Dubois, Li, Guestrin,
  Liang, and Hashimoto}]{alpaca}
Rohan Taori, Ishaan Gulrajani, Tianyi Zhang, Yann Dubois, Xuechen Li, Carlos
  Guestrin, Percy Liang, and Tatsunori~B. Hashimoto. 2023.
\newblock Stanford alpaca: An instruction-following llama model.
\newblock \url{https://github.com/tatsu-lab/stanford_alpaca}.

\bibitem[{Wang et~al.(2022)Wang, Mishra, Alipoormolabashi, Kordi, Mirzaei,
  Arunkumar, Ashok, Dhanasekaran, Naik, Stap, Pathak, Karamanolakis, Lai,
  Purohit, Mondal, Anderson, Kuznia, Doshi, Patel, Pal, Moradshahi, Parmar,
  Purohit, Varshney, Kaza, Verma, Puri, Karia, Sampat, Doshi, Mishra, Reddy,
  Patro, Dixit, Shen, Baral, Choi, Smith, Hajishirzi, and
  Khashabi}]{wang2022benchmarking}
Yizhong Wang, Swaroop Mishra, Pegah Alipoormolabashi, Yeganeh Kordi, Amirreza
  Mirzaei, Anjana Arunkumar, Arjun Ashok, Arut~Selvan Dhanasekaran, Atharva
  Naik, David Stap, Eshaan Pathak, Giannis Karamanolakis, Haizhi~Gary Lai,
  Ishan Purohit, Ishani Mondal, Jacob Anderson, Kirby Kuznia, Krima Doshi,
  Maitreya Patel, Kuntal~Kumar Pal, Mehrad Moradshahi, Mihir Parmar, Mirali
  Purohit, Neeraj Varshney, Phani~Rohitha Kaza, Pulkit Verma, Ravsehaj~Singh
  Puri, Rushang Karia, Shailaja~Keyur Sampat, Savan Doshi, Siddhartha Mishra,
  Sujan Reddy, Sumanta Patro, Tanay Dixit, Xudong Shen, Chitta Baral, Yejin
  Choi, Noah~A. Smith, Hannaneh Hajishirzi, and Daniel Khashabi. 2022.
\newblock \href {https://arxiv.org/abs/2204.07705} {Super-naturalinstructions:
  Generalization via declarative instructions on 1600+ tasks}.
\newblock In \emph{Conference on Empirical Methods in Natural Language
  Processing \CNFX{EMNLP}}.

\bibitem[{Wang et~al.(2021)Wang, Yu, Firat, and Cao}]{wang2021towards}
Zirui Wang, Adams~Wei Yu, Orhan Firat, and Yuan Cao. 2021.
\newblock \href {https://arxiv.org/abs/2109.09193} {Towards zero-label language
  learning}.
\newblock \emph{arXiv preprint arXiv:2109.09193}.

\bibitem[{Wei et~al.(2022)Wei, Bosma, Zhao, Guu, Yu, Lester, Du, Dai, and
  Le}]{wei2022finetuned}
Jason Wei, Maarten Bosma, Vincent Zhao, Kelvin Guu, Adams~Wei Yu, Brian Lester,
  Nan Du, Andrew~M. Dai, and Quoc~V Le. 2022.
\newblock \href {https://arxiv.org/abs/2109.01652} {{Finetuned Language Models
  are Zero-Shot Learners}}.
\newblock In \emph{International Conference on Learning Representations
  \CNFX{ICLR}}.

\bibitem[{Weir et~al.(2022)Weir, Yuan, C{\^o}t{\'e}, Hausknecht, Laroche,
  Momennejad, Van~Seijen, and Van~Durme}]{weir2022one}
Nathaniel Weir, Xingdi Yuan, Marc-Alexandre C{\^o}t{\'e}, Matthew Hausknecht,
  Romain Laroche, Ida Momennejad, Harm Van~Seijen, and Benjamin Van~Durme.
  2022.
\newblock \href {https://arxiv.org/abs/2203.04806} {{One-Shot Learning from a
  Demonstration with Hierarchical Latent Language}}.
\newblock \emph{arXiv preprint arXiv:2203.04806}.

\bibitem[{Welleck et~al.(2023)Welleck, Lu, West, Brahman, Shen, Khashabi, and
  Choi}]{welleck2022generating}
Sean Welleck, Ximing Lu, Peter West, Faeze Brahman, Tianxiao Shen, Daniel
  Khashabi, and Yejin Choi. 2023.
\newblock \href {https://arxiv.org/abs/2211.00053} {Generating sequences by
  learning to self-correct}.
\newblock In \emph{International Conference on Learning Representations
  \CNFX{ICLR}}.

\bibitem[{Weller et~al.(2020)Weller, Lourie, Gardner, and
  Peters}]{weller2020learning}
Orion Weller, Nicholas Lourie, Matt Gardner, and Matthew Peters. 2020.
\newblock \href {https://aclanthology.org/2020.emnlp-main.105/} {{Learning from
  Task Descriptions}}.
\newblock In \emph{Conference on Empirical Methods in Natural Language
  Processing \CNFX{EMNLP}}.

\bibitem[{West et~al.(2021)West, Bhagavatula, Hessel, Hwang, Jiang, Bras, Lu,
  Welleck, and Choi}]{west2021symbolic}
Peter West, Chandra Bhagavatula, Jack Hessel, Jena~D Hwang, Liwei Jiang,
  Ronan~Le Bras, Ximing Lu, Sean Welleck, and Yejin Choi. 2021.
\newblock \href {https://aclanthology.org/2022.naacl-main.341/} {Symbolic
  knowledge distillation: from general language models to commonsense models}.
\newblock In \emph{Conference of the North American Chapter of the Association
  for Computational Linguistics \CNFX{NAACL}}.

\bibitem[{Xie et~al.(2020)Xie, Luong, Hovy, and Le}]{xie2020self}
Qizhe Xie, Minh-Thang Luong, Eduard Hovy, and Quoc~V Le. 2020.
\newblock \href {https://arxiv.org/abs/1911.04252} {Self-training with noisy
  student improves imagenet classification}.
\newblock In \emph{IEEE Conference on Computer Vision and Pattern Recognition
  \CNFX{CVPR}}, pages 10687--10698.

\bibitem[{Xu et~al.(2023)Xu, Guo, Duan, and McAuley}]{xu2023baize}
Canwen Xu, Daya Guo, Nan Duan, and Julian McAuley. 2023.
\newblock \href {https://arxiv.org/abs/2304.01196} {Baize: An open-source chat
  model with parameter-efficient tuning on self-chat data}.
\newblock \emph{arXiv preprint arXiv:2304.01196}.

\bibitem[{Yang et~al.(2020)Yang, Malaviya, Fernandez, Swayamdipta, Bras, Wang,
  Bhagavatula, Choi, and Downey}]{yang2020generative}
Yiben Yang, Chaitanya Malaviya, Jared Fernandez, Swabha Swayamdipta, Ronan~Le
  Bras, Ji-Ping Wang, Chandra Bhagavatula, Yejin Choi, and Doug Downey. 2020.
\newblock \href {https://aclanthology.org/2020.findings-emnlp.90} {Generative
  data augmentation for commonsense reasoning}.
\newblock In \emph{Conference on Empirical Methods in Natural Language
  Processing \CNFX{EMNLP} - Findings}.

\bibitem[{Ye et~al.(2022)Ye, Kim, Jang, Shin, and Seo}]{ye2022guess}
Seonghyeon Ye, Doyoung Kim, Joel Jang, Joongbo Shin, and Minjoon Seo. 2022.
\newblock \href {https://arxiv.org/abs/2210.02969} {Guess the instruction!
  making language models stronger zero-shot learners}.
\newblock \emph{arXiv preprint arXiv:2210.02969}.

\bibitem[{Zelikman et~al.(2022)Zelikman, Mu, Goodman, and
  Wu}]{zelikman2022star}
Eric Zelikman, Jesse Mu, Noah~D Goodman, and Yuhuai~Tony Wu. 2022.
\newblock \href {https://arxiv.org/abs/2203.14465} {{ST}ar: Self-taught
  reasoner bootstrapping reasoning with reasoning}.
\newblock In \emph{Advances in Neural Information Processing Systems
  \CNFX{NeurIPS}}.

\bibitem[{Zhao et~al.(2022)Zhao, Ouyang, Yu, Wu, and Li}]{zhao2022pre}
Xuandong Zhao, Siqi Ouyang, Zhiguo Yu, Ming Wu, and Lei Li. 2022.
\newblock \href {https://arxiv.org/abs/2212.06950} {Pre-trained language models
  can be fully zero-shot learners}.
\newblock \emph{arXiv preprint arXiv:2212.06950}.

\bibitem[{Zhou et~al.(2022{\natexlab{a}})Zhou, He, Ma, Berg-Kirkpatrick, and
  Neubig}]{zhou2022prompt}
Chunting Zhou, Junxian He, Xuezhe Ma, Taylor Berg-Kirkpatrick, and Graham
  Neubig. 2022{\natexlab{a}}.
\newblock \href {https://arxiv.org/abs/2205.00049} {{Prompt Consistency for
  Zero-Shot Task Generalization}}.
\newblock In \emph{Conference on Empirical Methods in Natural Language
  Processing \CNFX{EMNLP} - Findings}.

\bibitem[{Zhou et~al.(2022{\natexlab{b}})Zhou, Muresanu, Han, Paster, Pitis,
  Chan, and Ba}]{zhou2022large}
Yongchao Zhou, Andrei~Ioan Muresanu, Ziwen Han, Keiran Paster, Silviu Pitis,
  Harris Chan, and Jimmy Ba. 2022{\natexlab{b}}.
\newblock \href {https://arxiv.org/abs/2211.01910} {Large language models are
  human-level prompt engineers}.
\newblock \emph{arXiv preprint arXiv:2211.01910}.

\end{thebibliography}
